\documentclass[a4paper,11pt]{article}
\usepackage{amsmath, amsthm, amssymb}
\usepackage{amsfonts}
\usepackage{ dsfont,bbold }
\usepackage{bm}
\usepackage{graphicx}
\usepackage{fullpage}
\usepackage{caption,subcaption}
\usepackage[colorlinks,linkcolor=blue,citecolor=blue,urlcolor=magenta,linktocpage,plainpages=false]{hyperref}\usepackage{color}

\usepackage{booktabs} 
\usepackage{pgfplots, pgfplotstable}
\pgfplotsset{compat=1.17}
\usepackage{tikz}
\usetikzlibrary{arrows,positioning}
\usetikzlibrary{external}
\usetikzlibrary{ calc, matrix, patterns, patterns.meta }
\usetikzlibrary{shapes,arrows, fit, tikzmark}
\usetikzlibrary{calc}
\usetikzlibrary{arrows}
\usetikzlibrary{decorations.pathreplacing}
\usepackage{multicol,multirow}

\usepgfplotslibrary{fillbetween}
\usepgfplotslibrary{statistics}
\definecolor{msdarkblue}{RGB}{36,58,94}
\definecolor{msblue}{RGB}{0,120,215}
\definecolor{msgreen}{RGB}{16,124,16}
\definecolor{msred}{RGB}{216,59,1}
\definecolor{msgray}{HTML}{DFDFDF}
\pgfplotsset{compat=1.6}

\newcommand{\circled}[1]{\tikz[baseline=(char.base)]{ \node[shape=circle,draw,inner sep=2pt] (char) {#1};}}

\usepackage{fontawesome5}
\usepackage{tcolorbox}
\usepackage{dashbox}

\tikzset{
    human icon/.pic={
        \node[draw=none] {\faUser};
    },
    robot icon/.pic={
        \node[draw=none ] (robotIcon) {\faRobot};

    }
}

\definecolor{msdarkblue}{RGB}{36,58,94}
\definecolor{msblue}{RGB}{0,120,215}
\definecolor{msgreen}{RGB}{16,124,16}
\definecolor{msred}{RGB}{216,59,1}
\definecolor{purple}{RGB}{128,0,128}
\definecolor{msgray}{HTML}{DFDFDF}
\definecolor{Emerald}{HTML}{00A99D}
\definecolor{RubineRed}{HTML}{ED017D}
\definecolor{pink}{HTML}{FF55A3}

\newtheoremstyle{mythmstyle}
{3pt}
{3pt}
{\color{red}}
{}
{\bfseries}
{.}
{.5em}
{}

\theoremstyle{mythmstyle}

\newcommand{\azure}{\textit{Azure}}
\newcommand{\name}{\textit{OptiGuide}}

\AtBeginDocument{%
  \providecommand\BibTeX{{%
    \normalfont B\kern-0.5em{\scshape i\kern-0.25em b}\kern-0.8em\TeX}}}

\usepackage{authblk}
\usepackage{spverbatim}

\newcommand{\QuestionItem}[1]{\tikz[baseline=(QuestionItem.base),remember
picture]{%
\node[fill=red!20,inner sep=4pt,font=\sffamily] (QuestionItem){#1};}}

\newcommand{\QuestionItemSecond}[1]{\tikz[baseline=(QuestionItemSecond.base),remember
picture]{%
\node[fill=red!20,inner sep=4pt,font=\sffamily] (QuestionItemSecond){#1};}}

\title{Large Language Models for Supply Chain Optimization}
\date{}
\author[1]{Beibin Li}
\author[1]{Konstantina Mellou}
\author[2]{Bo Zhang}
\author[2]{Jeevan Pathuri}
\author[1]{Ishai Menache}
\affil[1]{Microsoft Research}
\affil[2]{Microsoft Cloud Supply Chain}

\begin{document}
\maketitle

\renewcommand{\labelenumi}{\QuestionItem{Q\arabic{enumi}}}

\begin{abstract}
Supply chain operations traditionally involve a variety of complex decision making problems. Over the last few decades, supply chains greatly benefited from advances in computation, which allowed the transition from manual processing to automation and cost-effective optimization. Nonetheless, business operators still need to spend substantial efforts in \emph{explaining} and interpreting the optimization outcomes to stakeholders. Motivated by the recent advances in Large Language Models (LLMs), we study how this disruptive technology can help bridge the gap between supply chain automation and human comprehension and trust thereof. We design \name{} -- a framework that accepts as input queries in plain text, and outputs insights about the underlying optimization outcomes. Our framework does not forgo the state-of-the-art combinatorial optimization technology, but rather leverages it to quantitatively answer what-if scenarios (e.g., how would the cost change if we used supplier B instead of supplier A for a given demand?). Importantly, our design does not require sending proprietary data over to LLMs, which can be a privacy concern in some circumstances. We demonstrate the effectiveness of our framework on a real server placement scenario within Microsoft's cloud supply chain. Along the way, we develop a general evaluation benchmark, which can be used to evaluate the accuracy of the LLM output in other scenarios. 
\end{abstract}

\section{Introduction}

Modern supply chains are complex, containing multiple tiers of suppliers, customers, and service providers  \cite{hugos2018essentials}. Optimization tools have been widely utilized for decision making in such supply chains. These tools not only automate some of the decision making processes, but also result in efficiency gains and substantial cost reductions across many industries \cite{lambert2000issues}. However, some of the automated processes require involving business operators, for understanding and explaining certain decisions, providing what-if analysis, and even overriding some optimization outcomes. In many cases, these operators are not equipped with the necessary background in optimization, resulting in time-consuming back-and-forth interactions with program managers, data scientists and engineers.   

Large language models (LLMs) have recently emerged as a promising tool for assisting humans with a wide variety of tasks, such as writing documents, presenting work, coding and health diagnosis \cite{openai2023gpt4,bubeck2023sparks,lee2023benefits}. Generative multimodal LLMs, such as OpenAI's GPT-4, are being rapidly integrated within co-pilots, for answering questions and increasing productivity through simple, language based interactions with technology \cite{github_copilot}.

\begin{figure}
    \centering

\begin{tikzpicture}[
    human/.style={draw, fill=blue!30, text=black, rounded corners, align=left},
    bot/.style={draw, fill=green!40, text=black, rounded corners, align=left},
    thought/.style={draw, fill=gray!20, dashed, text=black, rounded corners, align=left},
    font=\sffamily
]

\node[human, anchor=north east] (h1) {Show me the shipping plan.};
\pic at ([xshift=-3mm, yshift=-2mm] h1.north west) {human icon};

\node[bot, anchor=north west, text width=10 cm] (b1) at ($(h1.south east) - (13 cm, 0.2 cm)$) {
    Here is the plan. You can hover your mouse on each node or edge to see more details. \\
    \begin{minipage}{3cm}
      \centering
       \begin{tikzpicture}[scale=0.5]
    \tikzstyle{arrow} = [thick,->,>=stealth, gray, text=black, font=\tiny]
    
    \node (S1) at (0,4) {S1};
    \node (S2) at (0,2) {S2};
    \node (S3) at (0,0) {S3};

    \node (R1) at (6,3) {R1};
    \node (R2) at (6,1) {R2};

    \node (C1) at (12,4) {C1};
    \node (C2) at (12,2) {C2};
    \node (C3) at (12,0) {C3};

    \draw [arrow] (S1) -- node[anchor=south]{80} (R2);

    \draw [arrow] (S2) -- node[anchor=east, yshift=2mm]{50} (R2);

    \draw [arrow] (S3) -- node[anchor=north,xshift=-4mm,yshift=-2mm]{100} (R1);

    \draw [arrow] (R1) -- node[anchor=south]{20 L + 20 D} (C1);
    \draw [arrow] (R1) -- node[anchor=west,yshift=1mm]{30 L + 20 D} (C2);
    \draw [arrow] (R1) -- node[anchor=south,yshift=-1mm,xshift=1mm]{10 L} (C3);

    \draw [arrow] (R2) -- node[anchor=north]{30 L + 100 D} (C3);
\end{tikzpicture}

    \end{minipage}
};
\pic at ([xshift=-3mm, yshift=-2mm] b1.north west) {robot icon};

\node[human, anchor=north east] (h3) at ($(b1.south -| h1.east) + (0, -2mm)$) {
Is it possible for Roastery 1 to be exclusively used by Cafe 2?
};
\pic at ([xshift=-3mm, yshift=-2mm] h3.north west) {human icon};

\node[bot, anchor=north west, text width=10 cm] (b3) at  ($(h3.south -| b1.west) + (0, -2mm)$) {
If we exclusively transport beans from Roastery 1 to Cafe 2, the cost would amount to \$2570, representing an approximate increase of $\frac{2570 - 2470}{2470} \approx 4\%$ compared to the current plan, which costs \$2470. Would you like to implement this change for future planning purposes?
};
\pic at ([xshift=-3mm, yshift=-2mm] b3.north west) {robot icon};

\node[draw=blue!70, rounded corners, anchor=north east, minimum width=13.5cm, minimum height=2mm, thick, text=gray!70] (message) at ($(b3.south -| h1.east) + (0, -2mm)$) {Type your question here...};

\node[draw, rounded corners, fit=(robotIcon) (h1) (b1)  (b3) (message) ] (window) {};

\end{tikzpicture}

\caption{\name{} interface. }
    \label{fig:chat_fig}
\end{figure}

In this paper, we study how state-of-the-art LLMs can be applied for reasoning about supply chain optimization. Using LLMs in our context is challenging. 
First, the underlying optimization problems are often large scale combinatorial optimization problems, and solving them directly is currently out of reach for LLMs \cite{bubeck2023sparks}. Second, one needs to align the large foundation models to answer the domain-specific questions. Due to the large scale, fully training these models is not possible, and even middle-ground solutions such as fine-tuning LLMs require substantial compute and engineering investments \cite{chen2023frugalgpt}. Last but not least, any use of LLMs in business-critical operations, should have solutions when ``things go wrong", including diagnosing of and recovering from mistakes and hallucinations \cite{liu2023prompt}.

In view of these challenges, we design and implement \name{} --  a framework that employs LLMs to interpret supply chain optimization solutions. A key idea behind \name{} is not to replace optimization technology by LLMs, but rather use optimization solvers in tandem with LLMs. In our design (see Figure \ref{fig:system} for system architecture), the LLM is responsible for translating the human query to ``optimization code", which is in turn used by an optimization solver to produce the necessary output; the output then passes through the LLM for producing the answer in human language (English). This architecture is used both for textual explanations and visualizations of the optimization solution, as well as for answering what-if queries.  To address what-if queries, \name{} uses the LLM to appropriately modify the input to the optimization solver, and then reruns the solver under the hood to produce an answer. 

To enable \name{}, we solve multiple technical challenges. First, we circumvent all forms of costly training, by applying in-context learning, namely ``teaching" the LLM about the domain directly through the query's prompt (i.e., as part of the inference). This requires careful co-design of the optimization code and the prompt with the understanding that the prompt can be space constrained. For example, we write the code in certain functional form that can be efficiently mapped to questions asked by humans.  
We also design a simple safeguard mechanism that confronts output mistakes. 

To evaluate the effectiveness of \name{}, we introduce an \emph{evaluation benchmark} that includes (i) a variety of common supply chain scenarios, and (ii) an evaluation methodology that incorporates new metrics for quantifying accuracy, generalizability within a scenario, and extrapolation capability to unseen scenarios. We test \name{} on five different scenarios and obtain 93\% accuracy on average using GPT-4. We view the benchmark and methodology as contributions that stand on their own, and can be used to evaluate future approaches. We are in the process of open-sourcing our benchmark. Finally, we deploy \name{} for the server deployment optimization used in Microsoft Azure's supply chain. We discuss some of the engineering challenges, and report initial promising results from our evaluation. 

We believe that this paper sets important foundations, which can be used by other organizations for explaining optimization outcomes through LLMs. There are several future directions that emerge from our study, for example, using smaller models that can be trained with modest resources. As a longer-term goal, it is natural to expand the scope of LLMs beyond explainability, to facilitate \emph{interactive} optimization (e.g., ``please provide a more load-balanced solution", ``please use at most two suppliers"). With the constant advances of LLM technology, it will be fascinating to examine whether LLMs can be utilized not only as translators, but also for refining and improving optimization outcomes. 

The rest of the paper is organized as follows. In Section \ref{sec:background}, we provide the necessary background on supply chain optimization and current LLM technology. In Section \ref{sec:framework}, we describe the design of \name{}.
Section \ref{sec:evaluation} describes our evaluation benchmark, and \name{}'s evaluation results. In Section \ref{sec:azure},
we outline our findings from \name{}'s deployment in Azure's supply chain. We discuss future perspectives in Section \ref{sec:discussion}.


\section{Background and Motivation} \label{sec:background}

In this section, we provide brief background on decision making in supply chain operations, and elaborate on the notion of explainability. We then describe current capabilities and limitations of LLMs, and conclude with a simple supply chain example, which will be useful for explaining our solution approach.

\begin{figure*}
    \centering
\tikzstyle{user} = [rectangle, rounded corners, minimum width=3cm, minimum height=1cm,text centered, draw=black, fill=red!30]
\tikzstyle{process} = [rectangle, minimum width=3cm, minimum height=1cm, text centered, draw=black, fill=orange!30]
\tikzstyle{decision} = [diamond, minimum width=3cm, minimum height=1cm, text centered, draw=black, fill=green!30]
\tikzstyle{box} = [rectangle, rounded corners, minimum width=3cm, minimum height=1cm, text centered, draw=black, fill=green!30]
\tikzstyle{tall_box} = [rectangle, rounded corners, minimum width=3cm, minimum height=1.2cm, text centered, draw=black, fill=green!30]
\tikzstyle{arrow} = [thick,->,>=stealth]
\tikzstyle{text_box} = [rectangle, minimum width=1mm, minimum height=1mm, text centered, draw=white]
\tikzstyle{narrow_box} = [rectangle, rounded corners, minimum width=2cm, minimum height=1cm, text centered, draw=black, fill=green!30]

\begin{tikzpicture}[node distance=1cm]

\node (user) [user] {{\Large \faUser} User};

\node (prompt) [tall_box, fill=blue!30, below=15mm of user] {Coder};
\node (safeguard) [tall_box, fill=blue!30, below=5mm of prompt] {Safeguard};
\node (interpret) [tall_box, fill=blue!30, below=5mm of safeguard] {Interpreter};
\node (text_opti) [text_box, below = 1mm of interpret.south] {\textit{Agents}};
\node (opti) [draw, rounded corners, dashed, dash pattern=on 2pt off 2pt, line width=1.5pt, fit= (prompt) (safeguard) (interpret)  (text_opti) ]  {};

\node (llm) [box, fill=yellow!30, right of=prompt, xshift=3.5cm, minimum height=2cm, minimum width=2cm] {LLM};

\node (t2) [narrow_box, left = 20mm of prompt.west, yshift=5mm] {Helper};
\node (t1) [narrow_box, left = 5mm of t2.west] {Solver};
\node (t3) [narrow_box, below = 3mm of t1.south] {Database};
\node (t4) [narrow_box, right = 5mm of t3.east] {Document};

\node (tools) [draw, rounded corners, dashed, fit=(t2) (t1) (t3) , dash pattern=on 2pt off 2pt, line width=1.5pt] {};
\node (text_tools) [text_box, above = 1mm of tools.north] {\textit{App Specific  Components}};

\draw [arrow] (user) -- (prompt) node[midway, right] { 
\circled{1} User question};
\draw [arrow] (prompt) -- node[anchor=north west, xshift=-8mm] {\circled{2} ICL} (llm);
\draw [arrow] ([xshift=-5mm] llm.south) |- node[anchor=north, xshift=-10mm] {\circled{3} Code} (safeguard.east);
\draw [arrow] (safeguard.west) -| ([xshift=10mm] tools.east)
node[anchor=south west, xshift=-11mm] {\circled{4} Input} 
|- (tools.east);
\draw [arrow] (tools.south) |- node[anchor=north west] {\circled{5} Output Logs} ([yshift=-3mm] opti.west);

\draw [arrow] ([yshift=5mm] interpret.east) -| node[anchor=north west, xshift=-20mm] {\circled{6} Result} ( llm.south);
\draw [arrow] ( [xshift=5mm] llm.south) |- node[anchor=north, xshift=-5mm] {\circled{7} Answer} ([yshift=-5mm] interpret.east);

\draw [arrow] (interpret.west) -| ([xshift=-3mm] tools.west) |- node[anchor=north west] {\circled{8} Final answer} (user.west); 

\end{tikzpicture}

    \caption{The \name{} framework.}
    \label{fig:system}
\end{figure*}

\subsection{Decision Making in Supply Chains}

A supply chain may be defined as ``an integrated network of facilities and transportation options for the supply, manufacture, storage, and distribution of materials and products'' \cite{garcia2015supply}. A simple supply chain may consist of a company (e.g., a service provider) and the set of its suppliers and customers \cite{hugos2018essentials}. However, most supply chains nowadays contain multiple tiers with suppliers of suppliers, customers of customers, and hierarchies of service providers \cite{hugos2018essentials}.
This results in highly complex global networks where decisions must be optimized across multiple layers to satisfy customer demand while guaranteeing operational efficiency.

Decision making in supply chains spans different time-scales: starting from the design of the supply chain network (e.g., location of factories), planning (e.g., procurement of supply), and execution (e.g., transportation of goods). This leads to many types of decisions; a few examples:
\begin{itemize}
    \itemsep0em
    \item How many factories should we open, where, and with what manufacturing capacity?
    \item What suppliers should we use?
    \item How much inventory should we keep in stock and at which locations?
    \item How should we transport intermediate and finished goods efficiently?
\end{itemize}

The complexity of the decision-making often requires the design of optimization approaches that can incorporate a multitude of constraints and objectives, and still generate good quality solutions in plausible running times. To this end, different aspects of the supply chain (facility location, inventory planning, routing) may be optimized separately or considered jointly (e.g., inventory planning integrated with routing \cite{pourhejazy2016new}). Common solution approaches for these optimization problems include Mixed Integer Programming based techniques and heuristics that can tackle the large scale of the problem.

\subsection{Explainability}

Business operators and planners involved in decision-making need to maintain a good understanding of the optimization outcomes. This allows them to not only address customer questions, but also react to unexpected events, and resolve inefficiencies and bottlenecks. However, the understanding is often challenging due to the complexity of the decision process (e.g., large scale, solution obtained by ``black-box" algorithm, etc.) and lack of optimization expertise. 

For concreteness, we provide below some examples of questions that operators may wish to answer.
\begin{enumerate}
    \itemsep0em
    \item What is the cost breakdown for each fulfilled demand?
    \item How much excess inventory have I had per month in the past year?
    \item What would happen if the demand at a particular location increased by $10\%$?
    \item Can I reduce a factory's manufacturing capacity by $5\%$ and still meet the demand?
    \item Why was a particular supplier selected for a demand?
    \item How would selecting a different transportation option affect the delivery timelines and the overall cost?
\end{enumerate}
These and other questions aim at \emph{explaining} the outcome of supply chain decisions. They include analyzing the current solution (input and output), investigating historical trends, and exploring what-if scenarios.

Obtaining insights on optimization decisions may require involving multiple professionals with different roles. Suppose that planners may wish to understand why a demand has not been fulfilled on time. They often surface the concern to the program managers, who involve domain experts, such as data scientists or the engineers that developed the optimization system. The domain experts in turn may need to write additional code and often rerun the optimization to extract the relevant insights. This overall process might be very time-consuming for all parties involved and can cause significant delays in the decision making process.

In some applications, teams maintain some custom tools that allow decision makers to reason about certain decisions. For example, application dashboards can provide visualizations or even allow enforcing some actions (e.g., fix a specific supplier for a demand). However, given the engineering overhead of maintaining 
the tools, they are typically limited to the most common use cases. 

The notion of explainability is certainly not novel, and has drawn attention in both academia and industry. There have been numerous studies on explaining ML/AI \cite{danilevsky2020survey,ahmed2022artificial}. In the optimization context, IBM Decision Optimization \cite{nickel2022decision} provides answers to a fixed set of queries that the user may choose to activate. See also \cite{vcyras2019argumentation} and references therein.

\subsection{Large Language Models}

\paragraph{Overview.}
A large language model (LLM) is a foundation model \cite{bommasani2021opportunities} trained on extensive text data using deep learning techniques, such as Transformer neural networks; ELMo \cite{peters2018deep}, BERT \cite{devlin2018bert}, Turing NLG \cite{rosset2020turing,smith2022using}, GPT-3 \cite{brown2020language}, GPT-4 \cite{openai2023gpt4}, PaLM \cite{chowdhery2022palm}, PaLM-E \cite{driess2023palm}, LLaMA \cite{touvron2023llama}, and Vicuna \cite{vicuna2023} are some examples of widely used LLMs.
In the training phase, a LLM learns statistical patterns, word relationships, and contextual information from diverse sources, such as books, articles, websites, and code repositories. LLMs are used for a variety of tasks in the inference phase \cite{bubeck2023sparks}, including chatbots, translation, writing assistance, coding  \cite{fried2022incoder,murali2023codecompose,kulal2019spoc}, planning \cite{liu2023llm+}, poem and story composition.

\paragraph{Using LLMs in applications.} Multiple strategies can be employed to adapt LLMs for a specific application.  The most common approaches are \emph{fine-tuning} and \emph{in-context learning}. 
Fine-tuning is a classic approach for ``transfer learning" aimed at transferring knowledge from a pre-trained LLM to a model tailored for a specific application \cite{weiss2016survey}. Typically, this process involves tweaking some weights of the LLM. While fine-tuning approaches can be made efficient \cite{lester2021power, dettmers2023qlora}, they still necessitate model hosting in GPUs. This requirement can prove excessively costly for many applications. In-context learning  \cite{dong2022survey} is an alternative cheaper approach, which involves incorporating a few training examples into the \emph{prompt} (or query). The idea here is to append the prompt with domain-specific examples and have the LLM learn from these ``few-shot" examples. A key advantage of this approach is that it does not require model parameter updates. 

\paragraph{Prompt engineering.} In a production setting, developers often send \emph{prompts} (aka, queries) to the model, which can be appended with domain-specific examples for obtaining higher-quality answers. A collection of prompt management tools, such as ChatGPT Plugin \cite{gptPlugin}, GPT function API call \cite{gptFuncCall}, LangChain  \cite{noauthor_langchain_nodate}, AutoGPT \cite{noauthor_auto-gpt_2023}, and BabyAGI \cite{noauthor_babyagi_nodate}, have been designed to help engineers integrate LLMs in applications and services. The prompt size is measured in the number of \emph{tokens}, which is proportional to the query size. LLMs can only process a limited number of tokens because of resource limitations, which is a strict constraint that developers and tools need to find workarounds for.

\paragraph{Privacy.} Using domain-specific information in the prompt may involve proprietary data, which users may prefer not to reveal to LLM hosts. Even if LLM providers offer service level agreements (SLAs) for privacy, passive eavesdropping attackers might still intercept the data. Therefore, many organizations would prefer utilizing LLMs in a privacy-preserving way, namely keeping the proprietary data in-house. 

\paragraph{Mistakes.} 
Naturally, LLMs might provide sub-optimal outcomes, such as inaccuracies and even hallucinations \cite{mckenna2023sources}. 
There are generic tools that tackle this problem \cite{manakul2023selfcheckgpt,peng2023check,sun2022contrastive}, however one may need domain specific tools for better outcomes. One example is fixing code generated by LLMs \cite{li2023skcoder,jiang2023self,wang2022compilable,chen2023improving}.

\subsection{A Simple Example} \label{sec:simple_example}

We now describe a simple supply chain example that will be useful for illustrating our approach. 

\begin{figure}
\centering
\begin{subfigure}[b]{0.56\textwidth}
    \centering
    \includegraphics[width=0.95\textwidth, page=2]{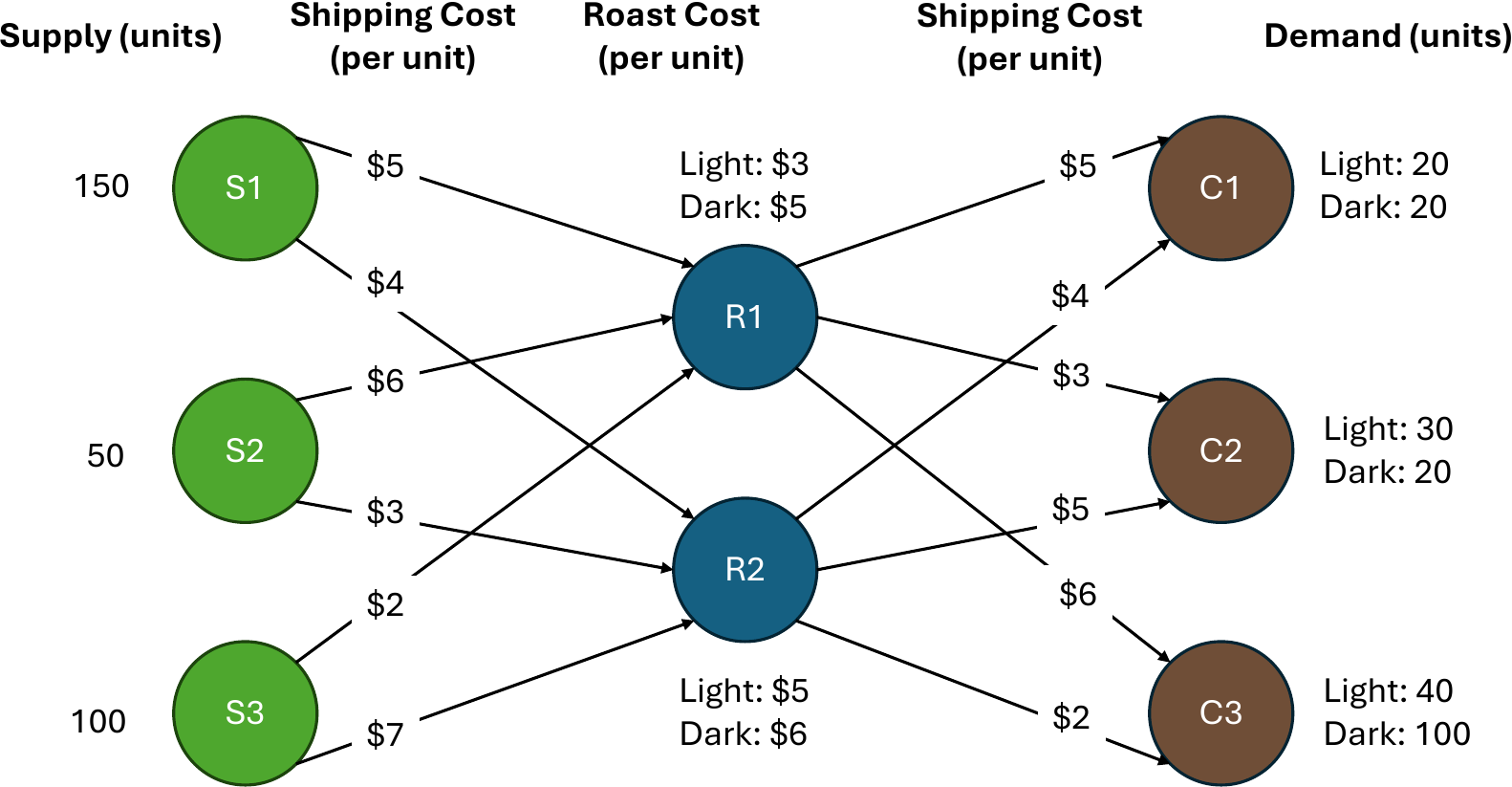}
    \caption{Problem setup.} 
\end{subfigure}
\begin{subfigure}[b]{0.4\textwidth}
    \centering
    \includegraphics[width=0.95\textwidth, page=3]{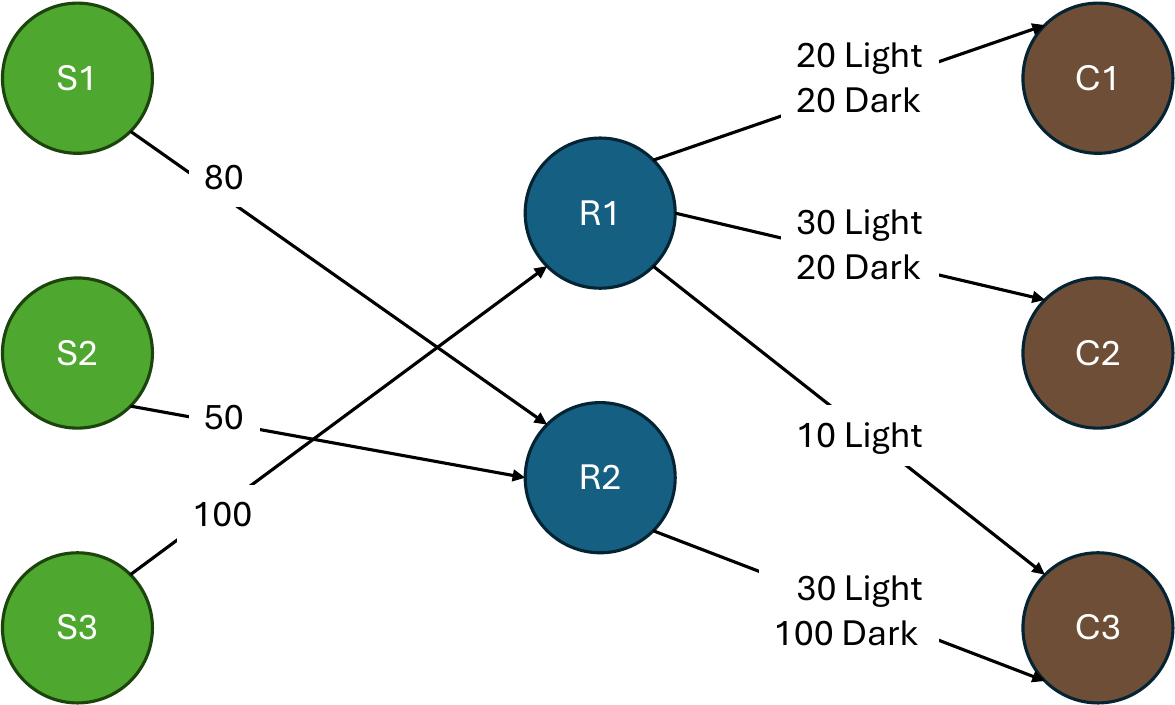}  
    \caption{Optimal plan (units).}  \label{fig:coffee-optimal-solution} 
    \end{subfigure}
    \caption{A simple supply chain example: coffee roasting company. } 
    \label{fig:cafe-problem}
\end{figure}

\paragraph{The supply chain.} Consider a coffee roasting company that roasts two types of coffee (light and dark roast). The company sources coffee beans from three different suppliers, it roasts them in one of its two roasting facilities, and then ships them to one of its three retail locations for selling to customers. The goal is to fulfill the demand in each retail location, while minimizing the total cost. The total cost consists of the cost of purchasing the coffee from the suppliers, the roasting cost in each facility, and the shipping cost of the end product to the retail locations. An illustration is given in Figure \ref{fig:cafe-problem}.

\paragraph{Model formulation.} We can model this problem as a Mixed Integer Program. Let $x_{s,r}$ denote the number of units purchased from supplier $s$ for roasting facility $r$, and $y^L_{r,\ell}$ and $y^D_{r,\ell}$ the amount of light and dark roast sent to retail location $\ell$ from roasting facility $r$. Each supplier $s$ has a capacity $C_s$, and each retail location $\ell$ has demand $D^L_{\ell}$ and $D^D_{\ell}$ for light and dark roast respectively.  There is a cost $c_{s,r}$ for each unit purchased from supplier $s$ for roasting facility $r$, a shipping cost of $g_{r,\ell}$ for each unit sent to retail location $\ell$ from roasting facility $r$, and a roasting cost $h_r^L$ and $h_r^D$ per unit of light roast and dark roast respectively in facility $r$. 
The optimization problem is the following:
\begin{align*}
\text{minimize} \quad  &   \Big( \sum_{s,r} x_{s,r} \cdot c_{s,r} +  
 \sum_{r,\ell}  y^L_{r,\ell} \cdot h^L_{r}+  & \\
& ~~~~ \sum_{r,\ell} y^D_{r,\ell} \cdot h^D_{r} + \sum_{r,\ell} (y^L_{r,\ell} + y^D_{r,\ell}) \cdot g_{r,\ell}  \Big)  \quad  & \text{(Objective)}  \\
\text{subject to} \quad & \sum_{r} x_{s,r} \leq C_{s} \quad & \forall s \quad \text{(Supplier capacity constraint)} \\
& \sum_{s} x_{s,r} = \sum_{\ell} (y^L_{r,\ell} + y^D_{r,\ell}) \quad & \forall r \quad \text{(Conservation of flow constraint)} \\
& \sum_{r} y^L_{r,\ell} \geq D^L_{\ell}  \quad & \forall \ell \quad \text{(Light coffee demand constraint)} \\
& \sum_{r} y^D_{r,\ell} \geq D^D_{\ell} \quad & \forall \ell \quad \text{(Dark coffee demand constraint)} \\
& x_{s,r}, y^L_{r,\ell}, y^D_{r,\ell} \in \mathbb{Z}^+ \quad & \forall s,r,\ell \quad \text{(Integrality constraint)}
\end{align*}

\renewcommand{\labelenumi}{\QuestionItemSecond{Q\arabic{enumi}}}

\paragraph{Explainability.} Let us now zoom into the example from Figure \ref{fig:cafe-problem}. The optimal solution is depicted in Figure \ref{fig:coffee-optimal-solution}. We see that in the optimal plan, both roasteries produce light and dark coffee; the first roastery sources its beans from supplier 3, while the second from suppliers 1 and 2. The first two retail locations then obtain all their coffee from the first roastery, while the third retail location is supplied by both roasteries. A user may ask the following questions:
\begin{enumerate}
    \itemsep0em 
    \item What would happen if the demand at retail location 1 increased by 10\%? 
    \item What would happen if the demands at all retail locations doubled? 
    \item Why are we using supplier 3 for roasting facility 1? 
    \item Can I use roasting facility 1 only for retail location 2? 
    \item What if supplier 3 can now provide only half of the quantity? 
    \item The per-unit cost from supplier 3 to roasting facility 1 is now \$5. How does that affect the total cost? 
    \item Why does Roastery 1 produce more light coffee than Roastery 2?
    \item Why does supplier 1 ship more to Roastery 2 than Roastery 1?
    \item Why not only use one supplier for Roastery 2?
\end{enumerate}


\section{The LLM Framework} \label{sec:framework}

Large-scale supply chain management entails multiple functions, such as extensive data gathering, data processing and analysis, optimization processes and communication and enforcement of decisions across multiple stakeholders. While LLMs and supporting tools may handle part of these functions, there is a need for an end-to-end framework that will address the underlying challenges in a systematic way. In this section, we describe the design of our framework, \name{}.

\subsection{System Overview}

The \name{} framework, depicted in Figure \ref{fig:system}, consists of three sets of entities: agents, LLMs, and application-specific components.  When a user poses a question (\circled{1}), the coder takes the question and formulates it as an in-context learning (ICL) question (\circled{2}) for the LLM. The LLM then generates code (\circled{3}) to answer the question. The safeguard checks the validity of the code and aborts the operation in case of a mistake; otherwise the safeguard feeds the code to an application specific component (\circled{4}), such as a database engine or an optimization solver (depending on the query). The component processes the code and produces results, which are logged in a file (\circled{5}). We note that obtaining the final result
may involve multiple iterations (\circled{2} to \circled{5}) where the query is automatically refined until the desired output is achieved. Finally, the output logs from the component are fed back into the LLM (\circled{6}). The LLM analyzes the logs and generates a human-readable answer (\circled{7}) that is sent back to the user (\circled{8}).
We now provide an overview of the different entities and components. More details can be found in Appendix \ref{sec:engineer_tricks}.

\subsubsection{Agents}

\begin{figure}[h]
    \includegraphics[width=0.95\textwidth, page=1]{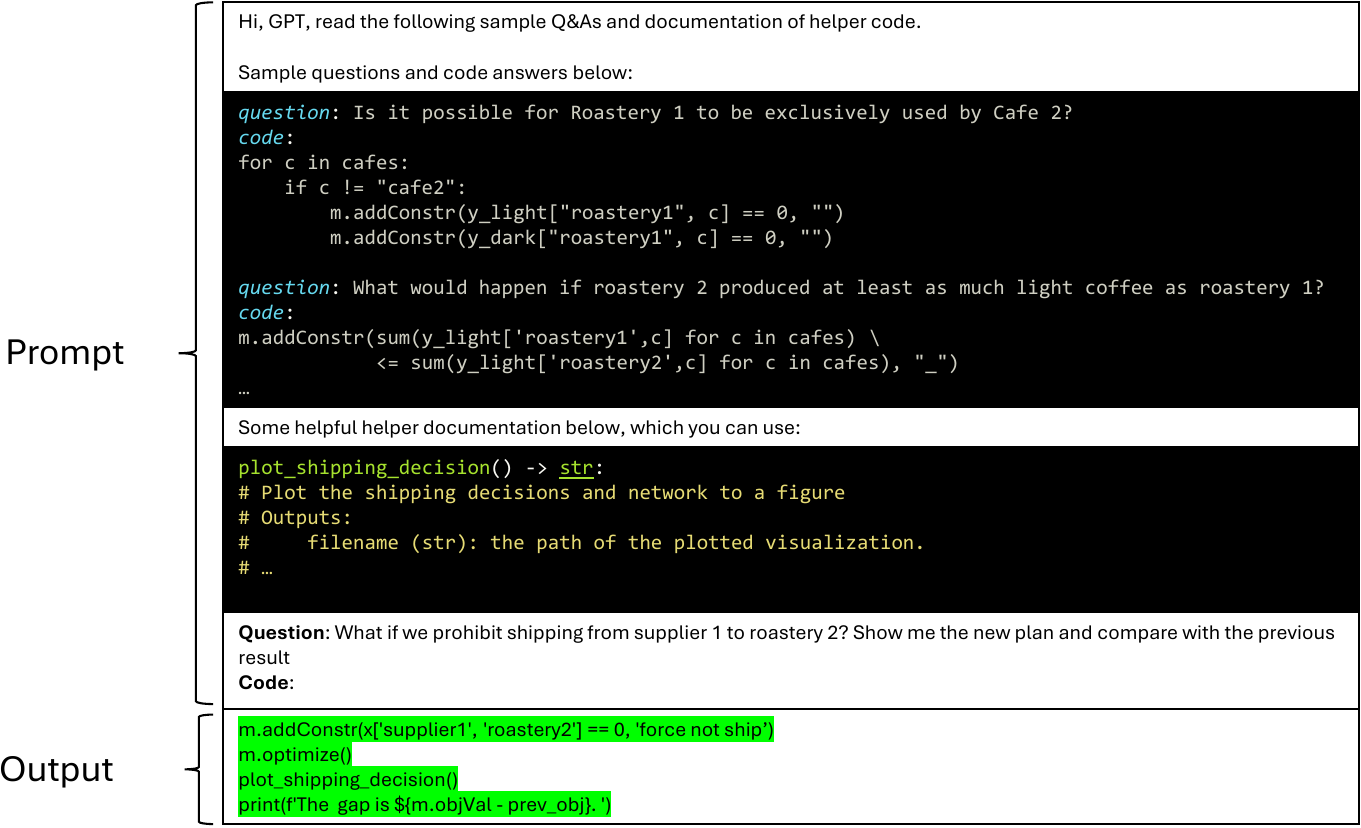}
    \caption{Coder prompt for the running example}
    \label{fig:icl}
\end{figure}

Agents facilitate the interaction between users, the LLM, and application-specific components. The \emph{coder} converts raw user questions into specific ICL queries. The conversion includes supplying the application context, providing ample training examples, and restructuring the user's query, as exemplified in Figure \ref{fig:icl}. The \emph{safeguard} operates as a quality control checkpoint. It scrutinizes the code for potential discrepancies and initiates self-debugging upon encountering failures. When \name{} cannot successfully address a query, the safeguard would either initiate a new iteration with a proposed fix, or generate an error message for the user. The \emph{interpreter} takes the output logs, tables, graphs, etc., and generates a human friendly response to the user's query. 

\subsubsection{Application Specific Components}

Different applications may have different types of components; we provide an overview of the most common ones. \name{} is designed in a modular way, so that using \name{} for a different application requires only switching to a new set of components.

The \emph{database} is a systematically arranged collection of data in various formats, such as CSV, SQL, JSON, Parquet, which are queried to extract answers. The \emph{solver} can be a commercial integer programming solver, such as Gurobi. \name{} can query the solver output directly, or the output can be stored and queried from the database. If a question demands profound domain knowledge or historical context, \name{} consults \emph{documents} to enhance the depth and relevance of the response. The \emph{helper} is an optional component. It consists of a set of functions written by application engineers, for simplifying the code produced by LLMs. For example, a complex data analysis workflow can be simplified to a single helper function call.

\begin{figure}[h]
    \includegraphics[width=0.95\textwidth, page=5]{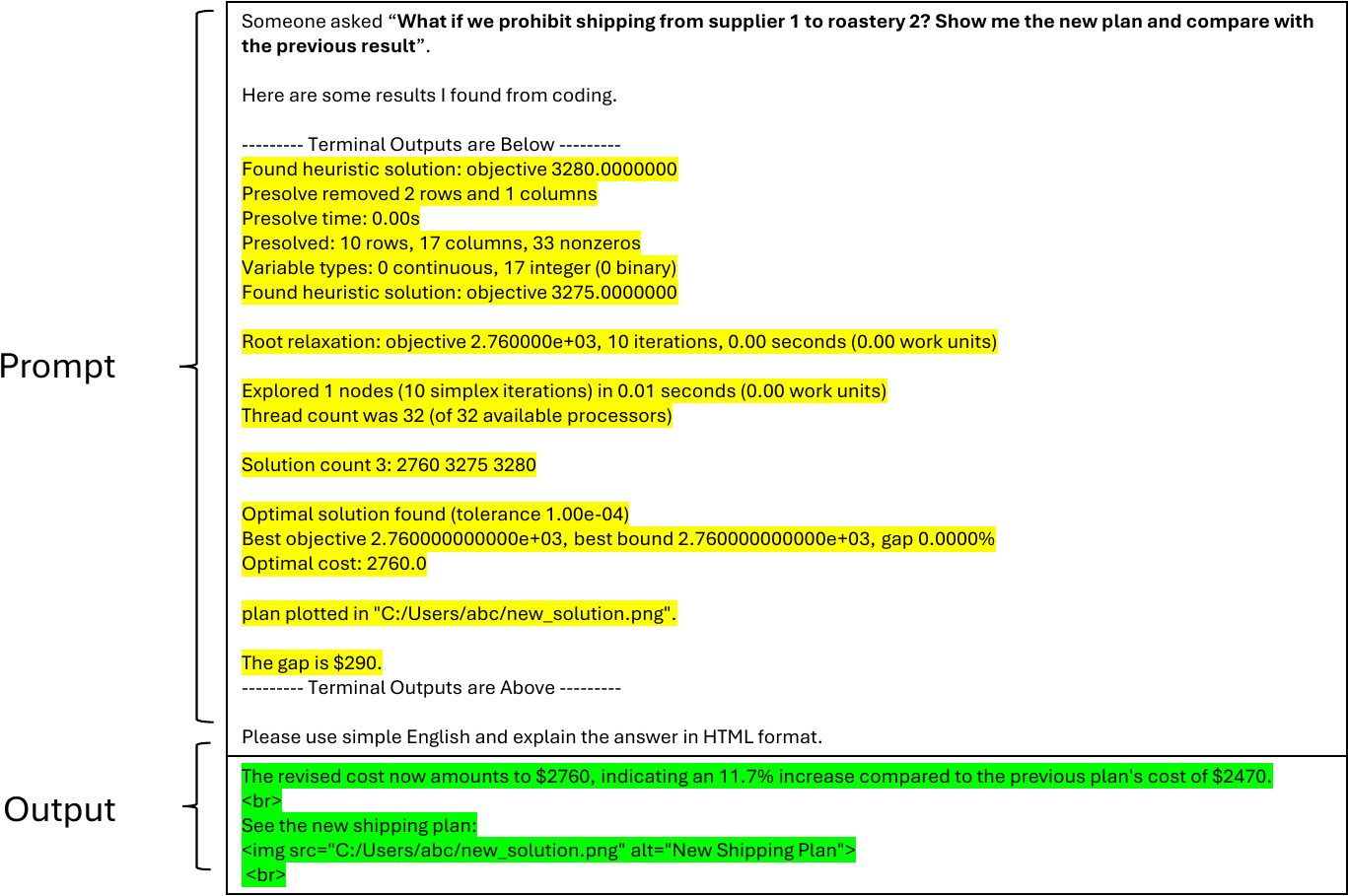}
    \caption{Interpreter prompt for the running example}
    \label{fig:interpreter}
\end{figure}

\subsection{A Running Example}

We illustrate \name{}'s data flow via the user question, ``What if we prohibit shipping from supplier 1 to roastery 2? Show me the new plan and compare with the previous result". First, the coder converts this question into an in-context learning query for the LLM, see Figure \ref{fig:icl} for the prompt. In addition to the question itself, the prompt contains (i) training examples, namely pairs of questions and code answers, and (ii) a documentation of the helper functions. Intuitively, (ii) supplements (i) by providing additional context into what the code does. 

Subsequently, the LLM generates code that adds a new constraint (green region in Figure \ref{fig:icl}). The safeguard then extracts the code from the LLM's response, and \name{} calls the optimization solver to resolve the planning problem, yielding a result depicted in the yellow region in Figure \ref{fig:interpreter}. This result is then fed into the LLM by the interpreter, which produces a response. Finally, \name{} presents the response to the user alongside a visualization of the plan (green region in Figure \ref{fig:interpreter}) and a comparison with the original cost. Note that \name{} preserves privacy, since the domain-specific data remains in either the solver or database, and is never transferred to the LLM. Additional examples are provided in Figure \ref{fig:detailed_example}.

\begin{figure}[h!]
    \centering

\begin{tikzpicture}[
    human/.style={draw, fill=blue!30, text=black, rounded corners, align=left},
    bot/.style={draw, fill=green!40, text=black, rounded corners, align=left},
    thought/.style={draw, fill=gray!20, dashed, text=black, rounded corners, align=left},
    font=\sffamily
]

\node[human, anchor=north east] (h1) {
Is it possible for Roastery 1 to be exclusively used by Cafe 2?
};
\pic at ([xshift=-3mm, yshift=-2mm] h1.north west) {human icon};

\node[thought, anchor=north west, text width=10 cm] (b1a) at ($(h1.south east) - (13 cm, 0.2 cm)$) {
We need to write Python code to test the what-if scneario. 
\begin{verbatim}
for c in cafes:
    if c != "cafe2":
        m.addConstr(
            y_light["roastery1", c] == 0, "")
        m.addConstr(
            y_dark["roastery1", c] == 0, "")
\end{verbatim}
};

\node[thought, anchor=north west, text width=10 cm, below=2mm of b1a.south] (b1b) {
The result has $m.objVal = \$2570$, which is higher than \$2470 from the previous solution.
};

\node[bot, anchor=north west, text width=10 cm, below=2mm of b1b.south] (b1c) {
If we exclusively transport beans from Roastery 1 to Cafe 2, the cost would amount to \$2570, representing an approximate increase of $\frac{2570 - 2470}{2470} \approx 4\%$ compared to the current plan, which costs \$2470. Would you like to implement this change for future planning purposes?
};
\pic at ([xshift=-3mm, yshift=-2mm] b1c.north west) {robot icon};

\node[human, anchor=north east] (h2) at ([yshift=-2mm] b1c.south -| h1.east) {
No. Who is the contact person for supplier 1?
};
\pic at ([xshift=-3mm, yshift=-2mm] h2.north west) {human icon};

\node[thought, anchor=north west, text width=10 cm] (b2a) at ($(h2.south east) - (13 cm, 0.2 cm)$) 
{
We can use SQL command to find out the information.
{\footnotesize
    \begin{verbatim}
SELECT contact FROM suppliers WHERE id="supplier1";
    \end{verbatim}
}

};

\node[thought, anchor=north west, text width=10 cm,  below=2mm of b2a.south] (b2b) { 
\textcolor{red}{\textbf{\large Safeguard Alert:} sensitive information. Approval required!}  \\
\textcolor{black}{Ticket submitted.}
};

\node[bot, anchor=north west, text width=10 cm,  below=2mm of b2b.south] (b2c) { To answer this question, we need approval from administrator to access privileged information. We just created a ticket and will get back to you in 12 hours... };
\pic at ([xshift=-3mm, yshift=-2mm] b2c.north west) {robot icon};

\node[draw=blue!70, rounded corners, anchor=north east, minimum width=13.5cm, minimum height=2mm, thick, text=gray!70] (message) at ($(b2c.south -| h1.east) + (0, -2mm)$) {Waiting for response...};

\node[draw, rounded corners, fit=(robotIcon) (h1) (b2c) (message) ] (window) {};

\end{tikzpicture}

\caption{
An illustration of questions answered by \name{}. 
The \colorbox{gray!20}{\begin{dashbox}{gray dashed boxes}\end{dashbox}} represent thoughts that occur in the backend. Users can configure \name{} to display these thoughts or not.
}

    \label{fig:detailed_example}
\end{figure}


\section{Evaluation Benchmark} \label{sec:evaluation}
\label{sec:eval}

In this section, we develop a benchmark for evaluating the performance of our framework on a variety of supply chain optimization problems. The benchmark and the methodology around it can guide future efforts for using LLMs in supply chain optimization.

\subsection{Scenarios and Data} \label{sec:eval_data}

To evaluate our framework, we selected a variety of optimization problems that capture multiple types of decisions that may be relevant in different supply chain settings. Specifically, our dataset includes a facility location scenario, a multi-commodity network flow for distribution of products, workforce assignment optimization, the traveling salesman problem, as well as the coffee distribution scenario from Section \ref{sec:simple_example}. The code for all problems is in Python and the Gurobi optimization solver \cite{bixby2007gurobi} is used to obtain the optimal solution; Appendix \ref{sec:coffee_code} provides the code for the coffee distribution problem as an example. 

Our next step is to generate a \emph{repository} of questions and code answers for each scenario. Some of these question-answer pairs will be used as examples for in-context learning, while others for evaluating \name{}'s performance. 
To create a large set of questions, we write macros for each question, which results in generating \emph{question sets} of closely related question-answer pairs. An example of a macro for a question set is the following:
\begin{samepage}
\begin{verbatim}
QUESTION: What if we prohibit shipping from {{VALUE-X}} to {{VALUE-Y}}?
VALUE-X: random.choice(suppliers)
VALUE-Y: random.choice(roasteries)
GROUND-TRUTH: model.addConstr(x[{{VALUE-X}}, {{VALUE-Y}}] == 0)
\end{verbatim}
\end{samepage}

In order to increase the diversity in the question sets, we also ask GPT to rephrase the questions while preserving their meaning. For instance, GPT might rephrase the generated question ``Why would we ship beans from Supplier 1 to Roastery 2'' to ``What benefits are associated with the choice of shipping beans from Supplier 1 to Roastery 2?''. 

We note that the question sets for all problems that are used in the benchmark were created from scratch and kept in house, so that the LLMs have not observed these data as part of their training.

\subsection{Evaluation Methodology}

The goal of our evaluation is to assess the accuracy of LLMs in answering user questions for supply chain optimization problems. Unfortunately, existing metrics, such as pass@k which is used for analyzing coding accuracy \cite{kulal2019spoc,chen2021evaluating}, are not well suited for explainability through code (intuitively, the metrics are ``too forgiving"). We therefore propose a different methodology which is inspired by the unit-test approach used in software development. 

Our evaluation proceeds as follows. For each scenario we run $R$ experiments. Each experiment consists of $T$ question sets. Each question set consists of $Q$ test questions and answers.
The LLM is asked to write the code and answer for a test question; it is given three chances to produce a response in case of an evident error (runtime or syntax). We then evaluate the correctness of the final answer. Note that we do not necessarily evaluate whether the generated code matches exactly with our ground-truth code, as there are different ways to obtain the correct response. The following example demonstrates a scenario where the generated code is quite different, but the optimization outcome would be the same.
\begin{verbatim}
1. model.addConstr(x['supplier1', 'roastery2'] == 0, 'force not ship')
2. shipping_cost_from_supplier_to_roastery[('supplier1', 'roastery2')] = 1e10
\end{verbatim}

\paragraph{Accuracy.} We define the accuracy metric $AC$ as the average success rate across all scenarios, experiments and question sets. Formally, \begin{align*}
    AC &= \frac{1}{  SR} \sum\limits_{s=1}^{S} \sum\limits_{r=1}^{R}  \frac{1}{T_s} \sum\limits_{t=1}^{T_s}  \mathds{1}(q_t),
\end{align*}
where $q_t$ is the question set, and $\mathds{1}(q_t)$ is the indicator whether it passed successfully. The LLM passes a question set if and only if it successfully answers all questions in the question set.

\begin{figure}
    \centering
   \includegraphics[width=1\textwidth, page=6]{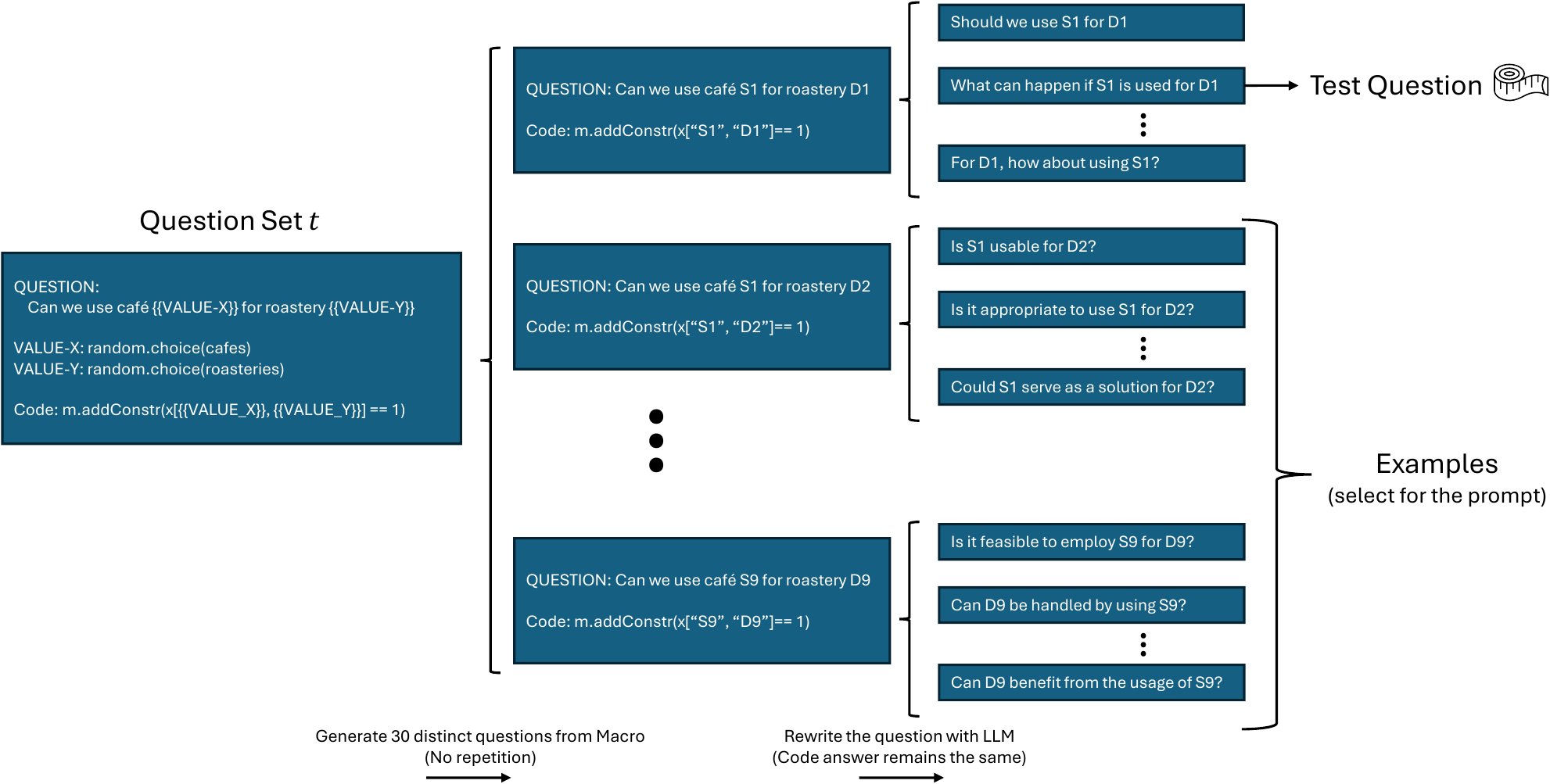}
    \caption{In-distribution evaluation}
    \label{fig:in-distribution}
\end{figure}

\begin{figure}
    \centering
   \includegraphics[width=0.5\textwidth, page=7]{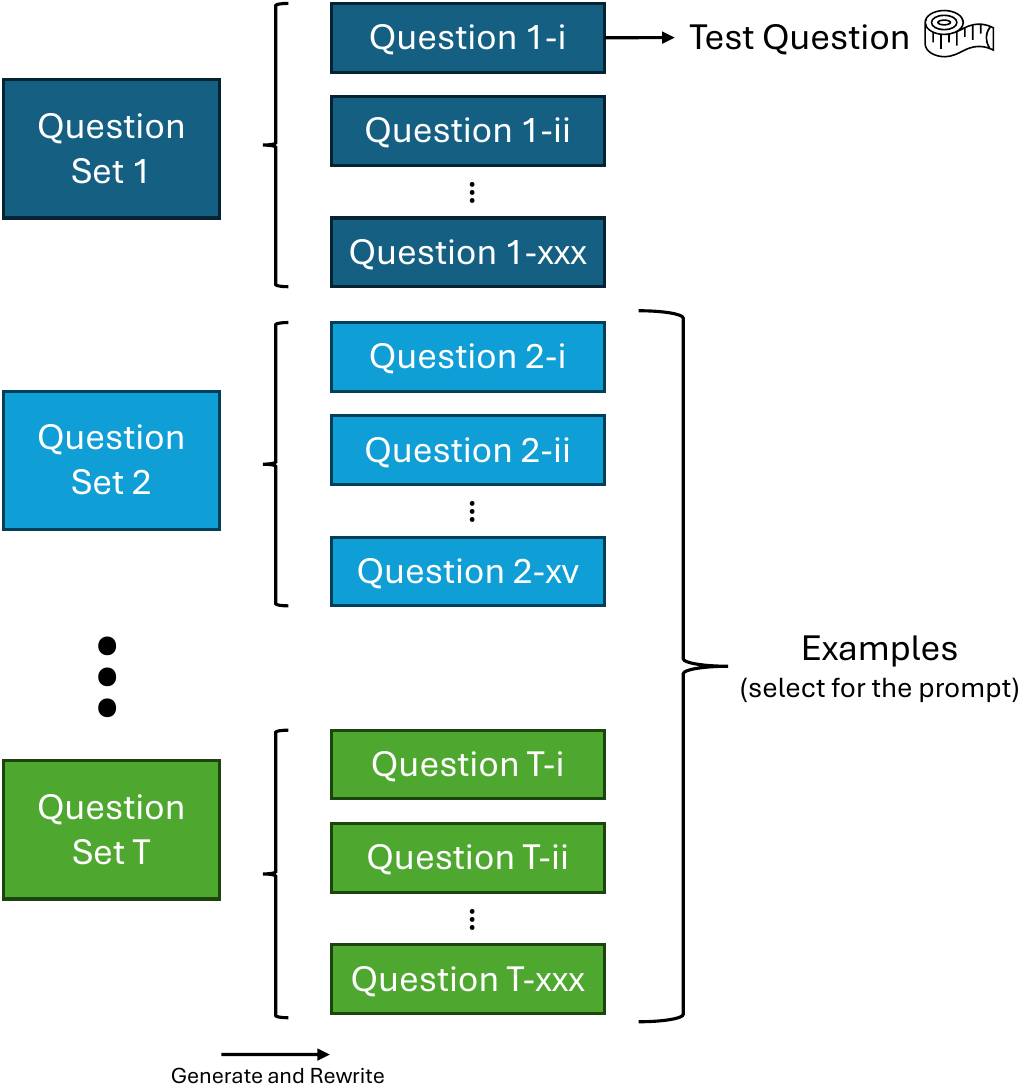}
    \caption{Out-of-distribution evaluation}
    \label{fig:out-of-distribution}
\end{figure}

\paragraph{In-distribution and out-of-distribution evaluation.} 
As common practice, we evaluate our framework in both `in-distribution' and `out-of-distribution'  \cite{zhou2022domain} settings.
For in-distribution evaluation (Figure \ref{fig:in-distribution}), the test question and the examples used in the prompt are from the same question set. In contrast, for out-of-distribution evaluation (Figure \ref{fig:out-of-distribution}), the example questions are extracted from different question sets.

\paragraph{Example selection.} As the number of tokens that can be provided as input to the LLMs is limited, we explore different approaches for selecting the training examples for each query. The approaches can be evaluated both for in-distribution and out-of-distribution evaluation. One approach is \emph{random selection}, where a fixed number of example questions is selected uniformly at random. Another approach is based on \emph{nearest neighbors}, where we select examples that are similar to the test question; similarity is based on the text embedding \cite{pennington2014glove} of the questions as determined by the model text-embedding-ada-002 \cite{brown2020language}. We also experiment with different sizes of the example set  (0, 1, 3, 5, or 10 examples).

\subsection{Performance}

\paragraph{Setup.} For each scenario $s$, we run $R=10$ experiments. In each experiment we evaluate $T_s\geq 10$ question sets. Each question set $q_t$ usually contains $10-30$ questions and answers.
We use both text-davinci-003 \cite{brown2020language} and GPT-4 \cite{openai2023gpt4} for our evaluation. 
Performance results across different LLMs, example selection approaches, and example set sizes are summarized in Table \ref{tab:perf}.

\begin{table}[]
\centering
\caption{Accuracy across different LLMs, example selection approaches, and example set sizes. Each experiment was run 10 times and the average accuracy is reported.}
\label{tab:perf}
\begin{tabular}{cc|cc|cc}
\toprule
                    &       & \multicolumn{2}{c}{In-distribution} & \multicolumn{2}{|c}{Out-of-distribution} \\
\# Examples              & Model & Random      & Nearest & Random       &  Nearest \\
\midrule
\multirow{2}{*}{0}	&	text-davinci-003	&	\multicolumn{4}{c}{0.32} \\
&	GPT-4	&	\multicolumn{4}{c}{0.59} \\
\midrule
\multirow{2}{*}{1}	&	text-davinci-003	&	0.78	&	0.78	&	0.39	&	0.44\\
&	GPT-4	&	0.85	&	0.90	&	0.66	&	0.66\\
\midrule
\multirow{2}{*}{3}	&	text-davinci-003	&	0.90	&	0.92	&	0.49	&	0.44\\
&	GPT-4	&	0.90	&	0.92	&	0.74	&	0.69\\
\midrule
\multirow{2}{*}{5}	&	text-davinci-003	&	0.93	&	0.93	&	0.52	&	0.48\\
&	GPT-4	&	0.92	&	0.93	&	0.78	&	0.73\\
\midrule
\multirow{2}{*}{10}	&	text-davinci-003	&	0.92	&	0.93	&	0.67	&	0.61\\
&	GPT-4	&	0.93	&	0.93	&	0.84	&	0.80\\
\bottomrule
\end{tabular}
\end{table}

\paragraph{Observations.} GPT-4 consistently outperforms text-davinci-003 in both in-distribution and out-of-distribution evaluation. 
As expected, both models show higher accuracy on in-distribution  compared to out-of-distribution evaluation. GPT-4 performs relatively much better in out-of-distribution evaluation, demonstrating its  stronger reasoning and generalization capabilities; another sign for these capabilities is the 59\% accuracy even without any training examples. Increasing the number of examples results in improved accuracy across the board. We also note that the gap between text-davinci-003 and GPT-4 decreases with the size of the example set. 

The nearest neighbor selection approach yields slight performance improvements for in-distribution evaluation. Interestingly, when the size of the example set is greater than one, random selection outperforms nearest neighbor for out-of-distribution evaluation. One explanation here is that selecting examples based on text similarity results in overfitting, and random selection results in more diverse training examples.


\section{\name{} for Azure's Supply Chain} \label{sec:azure}
\label{sec:ifs}

In this section, we demonstrate \name{}'s capabilities on the server fulfillment supply chain of Microsoft Azure. We start with providing the necessary details for the decisions involved in Azure's supply chain. We then outline the steps for deploying \name{} in production, and provide examples of user interactions and early feedback we obtained. We conclude this section by describing preliminary performance results.

\begin{figure}[h!]
    \centering
    \includegraphics[width=0.9\textwidth, page=4]{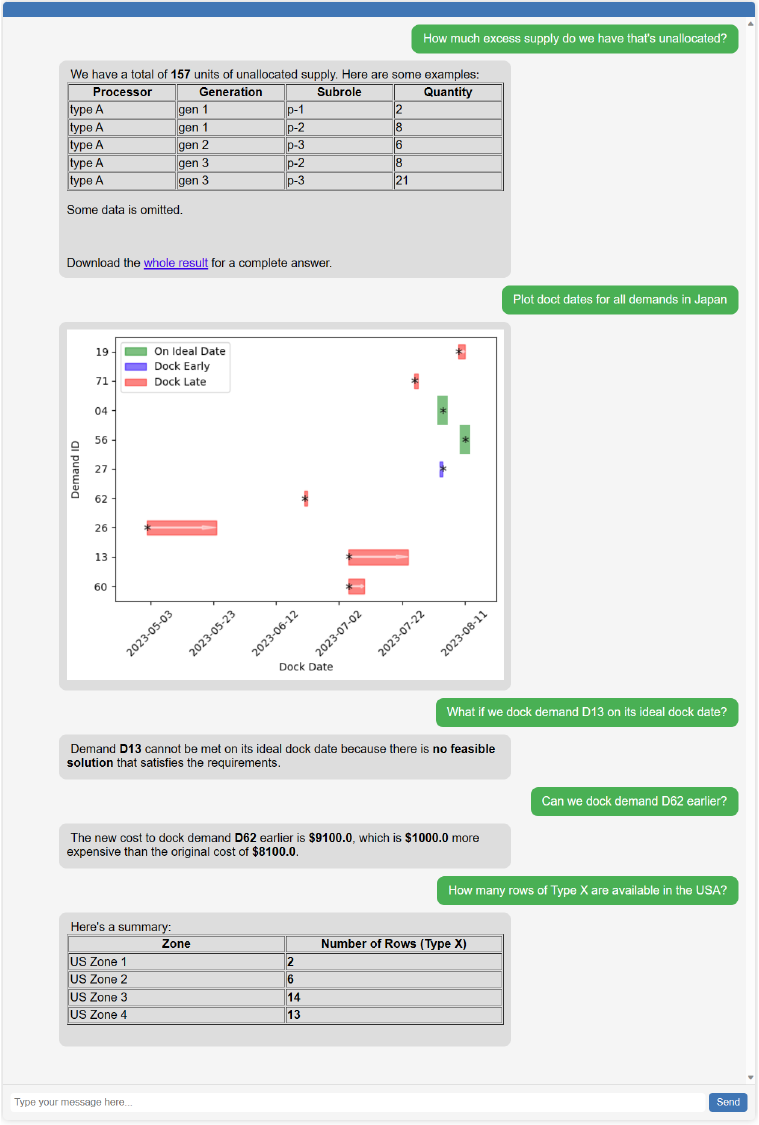}
    \caption{Screenshot of \name{} in Microsoft \azure{} production. We anonymized names and data by using generic values.}
    \label{fig:screenshots}
\end{figure}

\subsection{The \azure{} Supply Chain}
The rapid growth of the cloud industry requires cloud providers to continuously deploy additional capacity to keep up with the demand. This is achieved by acquiring new clusters of servers and deploying them in the data centers. The Microsoft Azure supply chain encompasses a broad array of processes including demand forecasting, strategic foresight, hardware semantic search, fulfillment planning, and document management. Due to complexity and large scale, the optimization of Azure's supply chain is assigned to different subsystems. We focus here on one such subsystem called Intelligent Fulfillment System (IFS), which deals with assigning and shipping servers from the warehouse to the data centers. 

\paragraph{Main decisions.} For each demand for cloud capacity, the main decisions consist of (i) the hardware supplier that will be used to fulfill the demand, (ii) the timeline of the deployment - in particular, the cluster's dock-date (which determines the date of shipping from the warehouse), and (iii) the cluster's deployment location in the data center (selection of a row of tiles to place the cluster on). The goal is to minimize the total cost that consists of multiple components, such as delay/idle cost of the clusters compared to their ideal dock-date and shipping costs, while respecting a multitude of constraints.  Examples of constraints include capacity constraints on the suppliers and the data centers, location preferences for demands and compatibility constraints. The underlying optimization problem is formulated as a Mixed Integer Program (MIP) where the total input data size is around 500 MB.
The optimal solution is obtained hourly using Gurobi. More details about the optimization problem can be found in Appendix \ref{sec:app:ifs}. 

\paragraph{Stakeholders.} The main consumers of IFS are \emph{planners}. These are professionals that have the business context, so when they receive the outcome of the optimization, they can confirm that it meets business needs (or override decisions otherwise) and ensure the execution of the decisions is completed as planned. However, the increased complexity of the underlying optimization problem in combination with the global scale of decision making (hundreds of data centers) prevents immediate clarity in the reasoning behind each decision. Consequently, planners often reach out to the \emph{engineers} (including data scientists) that develop the optimization system for obtaining additional insights. 
Oftentimes, planners and engineers have multiple rounds of interaction around understanding an issue or exploring what-if scenarios. 

\paragraph{Common questions.} We summarize below the main types of questions that are raised by planners:

\begin{enumerate}
    \itemsep0em 
    \item {[}Management{]} Does the system support a particular region, resource, or supplier?
    \item {[}Availability{]} Is a resource available or allocated?
    \item {[}Decisions{]} Why did the system make decision `x' related to supplier/demand selection, time, and location?
    \item {[}Details of shipments{]} What are the details related to cross-geographical shipments and expected dock counts on a specific date?
    \item {[}Historical data analysis{]} What is the standard deviation of the supplier's inventory in the last month?
    \item {[}Visualization{]} Can you visualize the dock capacity, availability, dates, or delays at a given location?
\end{enumerate}

\subsection{Deploying \name{} for Azure Supply Chain}

Our current deployment of \name{} consists of (i) a front-end service for multiple-user interaction; (ii) an agent service, which is connected to Azure OpenAI for LLM access; (iii) multiple virtual machines (VMs) which host IFS and the application specific components to support multiple users at the same time.

We preload VMs' memories with the input data and solver's solutions to speedup code executions for users. The input data for the optimization problem are updated periodically (hourly), where the VMs load the updated data in a round-robin fashion so that there are always some VMs available to support users. We use GPT-4 as the LLM.

\subsection{Preliminary Feedback and Results}

Figure \ref{fig:screenshots} provides examples of interactions between users and \name{}.

The preliminary feedback we obtained from both planners and engineers has been positive. Users expressed excitement noting the potential of \name{} to help them understand the underlying optimization logic. Users especially emphasized the benefits of supporting key what-if scenarios, which gives planners more autonomy and may substantially reduce the engineering on-call burden. For example, before \name{}, answering one what-if question would need more than three operators to coordinate the investigation and one on-call engineer to inspect the plan output.

Our preliminary evaluation indicates that \name{} can achieve more than 90\% accuracy for our in-distribution evaluation. This result is consistent with the ones obtained in Section \ref{sec:eval}.


\section{Concluding Remarks} \label{sec:discussion}
We conclude this paper by discussing current limitations, and highlighting intriguing directions for future work. 

\subsection{Current Limitations}

\paragraph{Users need to be specific.} The user needs to ask precise questions. For instance, ``Can we dock demand xc132 fifteen days earlier?" is ambiguous, because ``earlier" can mean ``15 days before today", ``15 days before the currently planned date", or ``15 days before the deadline". Consequently, the LLM might misunderstand the user and yield the wrong code. 

\paragraph{Dependency on application-specific components.} 
\name{} relies on proper design of application-specific components, such as the schema of the database and the helper functions. Some of these components might require non-negligible engineering efforts. While there has been progress in automating some of these components \cite{cai2023large}, there are still gaps in using them in some production settings.

\paragraph{Undetected mistakes.} We observed cases where the LLM writes code that runs smoothly, but it may be totally wrong (e.g., due to string matching mistakes). We expect that things will improve in the future with more advances in LLMs and supporting tools. 

\paragraph{Generalize to new questions.}  While the LLM performs well on seen questions, it still struggles when presented with questions that do not appear in the examples (see, e.g., Table \ref{tab:perf}). We believe that future models will have better generalizability.

\paragraph{Benchmark.} Our current evaluation quantifies performance only for quantitative questions; for example, we exclude visualization queries from our analysis. Furthermore, the evaluation is based on a specific programming language (Python) and optimization solver (Gurobi).

\subsection{Future Directions}

We see our work as a cornerstone for future research in the area. 
One interesting direction is incorporating human feedback (e.g., from supply chain planners) which could lead to significant performance improvements \cite{wiegreffe2021reframing}. Another direction that we are currently examining is using smaller models (see, e.g., \cite{gunasekar2023textbooks} and references therein) for the specific tasks of supply chain optimization; using such models allows for more affordable hosting and fine-tuning of the model. In particular, we are examining whether fine-tuning can help with interpreting unseen questions. On a related note, it is of interest to consider a hybrid framework that combines the strengths of different AI models, for example combining large LMs with smaller ones.  A natural longer-term goal is to go beyond explainability and facilitate \emph{interactive} optimization, where the user directly influences the optimization outcomes; this will require designing more comprehensive safeguards, to prevent costly mistakes.

\subsection*{Acknowledgements}
\begin{samepage}
We thank S\'ebastien Bubeck, Yin Tat Lee, Chi Wang, Erkang Zhu, Leonardo Nunes, Srikanth Kandula, Adam Kalai, Marco Molinaro, Luke Marshall, Patricia Kovaleski, Hugo Barbalho, Tamires Santos, Runlong Zhou, Ashley Llorens, Surajit Chaudhuri, and Johannes Gehrke from Microsoft Research for useful discussions. We also thank Brian Houser, Matthew Meyer, Ryan Murphy, Russell Borja, Yu Ang Zhang,  Rojesh Punnath, Naga Krothapalli, Navaneeth Echambadi, Apoorav Trehan, Jodi Larson, and Cliff Henson from the Microsoft Cloud Supply Chain for their advice and support.
\end{samepage}

\bibliography{ref.bib}

\begin{thebibliography}{10}

\bibitem{hugos2018essentials}
Michael~H Hugos.
\newblock {\em Essentials of supply chain management}.
\newblock John Wiley \& Sons, 2018.

\bibitem{lambert2000issues}
Douglas~M Lambert and Martha~C Cooper.
\newblock Issues in supply chain management.
\newblock {\em Industrial marketing management}, 29(1):65--83, 2000.

\bibitem{openai2023gpt4}
OpenAI.
\newblock Gpt-4 technical report, 2023.

\bibitem{bubeck2023sparks}
S{\'e}bastien Bubeck, Varun Chandrasekaran, Ronen Eldan, Johannes Gehrke, Eric
  Horvitz, Ece Kamar, Peter Lee, Yin~Tat Lee, Yuanzhi Li, Scott Lundberg,
  et~al.
\newblock Sparks of artificial general intelligence: Early experiments with
  gpt-4.
\newblock {\em arXiv preprint arXiv:2303.12712}, 2023.

\bibitem{lee2023benefits}
Peter Lee, Sebastien Bubeck, and Joseph Petro.
\newblock Benefits, limits, and risks of gpt-4 as an ai chatbot for medicine.
\newblock {\em New England Journal of Medicine}, 388(13):1233--1239, 2023.

\bibitem{github_copilot}
GitHub.
\newblock Github copilot: Your ai pair programmer, 2023.

\bibitem{chen2023frugalgpt}
Lingjiao Chen, Matei Zaharia, and James Zou.
\newblock Frugalgpt: How to use large language models while reducing cost and
  improving performance.
\newblock {\em arXiv preprint arXiv:2305.05176}, 2023.

\bibitem{liu2023prompt}
Yi~Liu, Gelei Deng, Yuekang Li, Kailong Wang, Tianwei Zhang, Yepang Liu, Haoyu
  Wang, Yan Zheng, and Yang Liu.
\newblock Prompt injection attack against llm-integrated applications.
\newblock {\em arXiv preprint arXiv:2306.05499}, 2023.

\bibitem{garcia2015supply}
Daniel~J Garcia and Fengqi You.
\newblock Supply chain design and optimization: Challenges and opportunities.
\newblock {\em Computers \& Chemical Engineering}, 81:153--170, 2015.

\bibitem{pourhejazy2016new}
Pourya Pourhejazy and Oh~Kyoung Kwon.
\newblock The new generation of operations research methods in supply chain
  optimization: A review.
\newblock {\em Sustainability}, 8(10):1033, 2016.

\bibitem{danilevsky2020survey}
Marina Danilevsky, Kun Qian, Ranit Aharonov, Yannis Katsis, Ban Kawas, and
  Prithviraj Sen.
\newblock A survey of the state of explainable ai for natural language
  processing.
\newblock {\em arXiv preprint arXiv:2010.00711}, 2020.

\bibitem{ahmed2022artificial}
Imran Ahmed, Gwanggil Jeon, and Francesco Piccialli.
\newblock From artificial intelligence to explainable artificial intelligence
  in industry 4.0: a survey on what, how, and where.
\newblock {\em IEEE Transactions on Industrial Informatics}, 18(8):5031--5042,
  2022.

\bibitem{nickel2022decision}
Stefan Nickel, Claudius Steinhardt, Hans Schlenker, and Wolfgang Burkart.
\newblock {\em Decision Optimization with IBM ILOG CPLEX Optimization Studio: A
  Hands-On Introduction to Modeling with the Optimization Programming Language
  (OPL)}.
\newblock Springer Nature, 2022.

\bibitem{vcyras2019argumentation}
Kristijonas {\v{C}}yras, Dimitrios Letsios, Ruth Misener, and Francesca Toni.
\newblock Argumentation for explainable scheduling.
\newblock In {\em Proceedings of the AAAI Conference on Artificial
  Intelligence}, volume~33, pages 2752--2759, 2019.

\bibitem{bommasani2021opportunities}
Rishi Bommasani, Drew~A Hudson, Ehsan Adeli, Russ Altman, Simran Arora, Sydney
  von Arx, Michael~S Bernstein, Jeannette Bohg, Antoine Bosselut, Emma
  Brunskill, et~al.
\newblock On the opportunities and risks of foundation models.
\newblock {\em arXiv preprint arXiv:2108.07258}, 2021.

\bibitem{peters2018deep}
Matthew~E. Peters, Mark Neumann, Mohit Iyyer, Matt Gardner, Christopher Clark,
  Kenton Lee, and Luke Zettlemoyer.
\newblock Deep contextualized word representations, 2018.

\bibitem{devlin2018bert}
Jacob Devlin, Ming-Wei Chang, Kenton Lee, and Kristina Toutanova.
\newblock Bert: Pre-training of deep bidirectional transformers for language
  understanding.
\newblock {\em arXiv preprint arXiv:1810.04805}, 2018.

\bibitem{rosset2020turing}
Corby Rosset.
\newblock Turing-nlg: A 17-billion-parameter language model by microsoft.
\newblock {\em Microsoft Blog}, 1(2), 2020.

\bibitem{smith2022using}
Shaden Smith, Mostofa Patwary, Brandon Norick, Patrick LeGresley, Samyam
  Rajbhandari, Jared Casper, Zhun Liu, Shrimai Prabhumoye, George Zerveas,
  Vijay Korthikanti, et~al.
\newblock Using deepspeed and megatron to train megatron-turing nlg 530b, a
  large-scale generative language model.
\newblock {\em arXiv preprint arXiv:2201.11990}, 2022.

\bibitem{brown2020language}
Tom Brown, Benjamin Mann, Nick Ryder, Melanie Subbiah, Jared~D Kaplan, Prafulla
  Dhariwal, Arvind Neelakantan, Pranav Shyam, Girish Sastry, Amanda Askell,
  et~al.
\newblock Language models are few-shot learners.
\newblock {\em Advances in neural information processing systems},
  33:1877--1901, 2020.

\bibitem{chowdhery2022palm}
Aakanksha Chowdhery, Sharan Narang, Jacob Devlin, Maarten Bosma, Gaurav Mishra,
  Adam Roberts, Paul Barham, Hyung~Won Chung, Charles Sutton, Sebastian
  Gehrmann, Parker Schuh, Kensen Shi, Sasha Tsvyashchenko, Joshua Maynez,
  Abhishek Rao, Parker Barnes, Yi~Tay, Noam Shazeer, Vinodkumar Prabhakaran,
  Emily Reif, Nan Du, Ben Hutchinson, Reiner Pope, James Bradbury, Jacob
  Austin, Michael Isard, Guy Gur-Ari, Pengcheng Yin, Toju Duke, Anselm
  Levskaya, Sanjay Ghemawat, Sunipa Dev, Henryk Michalewski, Xavier Garcia,
  Vedant Misra, Kevin Robinson, Liam Fedus, Denny Zhou, Daphne Ippolito, David
  Luan, Hyeontaek Lim, Barret Zoph, Alexander Spiridonov, Ryan Sepassi, David
  Dohan, Shivani Agrawal, Mark Omernick, Andrew~M. Dai,
  Thanumalayan~Sankaranarayana Pillai, Marie Pellat, Aitor Lewkowycz, Erica
  Moreira, Rewon Child, Oleksandr Polozov, Katherine Lee, Zongwei Zhou, Xuezhi
  Wang, Brennan Saeta, Mark Diaz, Orhan Firat, Michele Catasta, Jason Wei,
  Kathy Meier-Hellstern, Douglas Eck, Jeff Dean, Slav Petrov, and Noah Fiedel.
\newblock Palm: Scaling language modeling with pathways, 2022.

\bibitem{driess2023palm}
Danny Driess, Fei Xia, Mehdi~SM Sajjadi, Corey Lynch, Aakanksha Chowdhery,
  Brian Ichter, Ayzaan Wahid, Jonathan Tompson, Quan Vuong, Tianhe Yu, et~al.
\newblock Palm-e: An embodied multimodal language model.
\newblock {\em arXiv preprint arXiv:2303.03378}, 2023.

\bibitem{touvron2023llama}
Hugo Touvron, Thibaut Lavril, Gautier Izacard, Xavier Martinet, Marie-Anne
  Lachaux, Timoth{\'e}e Lacroix, Baptiste Rozi{\`e}re, Naman Goyal, Eric
  Hambro, Faisal Azhar, et~al.
\newblock Llama: Open and efficient foundation language models.
\newblock {\em arXiv preprint arXiv:2302.13971}, 2023.

\bibitem{vicuna2023}
Wei-Lin Chiang, Zhuohan Li, Zi~Lin, Ying Sheng, Zhanghao Wu, Hao Zhang, Lianmin
  Zheng, Siyuan Zhuang, Yonghao Zhuang, Joseph~E. Gonzalez, Ion Stoica, and
  Eric~P. Xing.
\newblock Vicuna: An open-source chatbot impressing gpt-4 with 90\%* chatgpt
  quality, March 2023.

\bibitem{fried2022incoder}
Daniel Fried, Armen Aghajanyan, Jessy Lin, Sida Wang, Eric Wallace, Freda Shi,
  Ruiqi Zhong, Wen-tau Yih, Luke Zettlemoyer, and Mike Lewis.
\newblock Incoder: A generative model for code infilling and synthesis.
\newblock {\em arXiv preprint arXiv:2204.05999}, 2022.

\bibitem{murali2023codecompose}
Vijayaraghavan Murali, Chandra Maddila, Imad Ahmad, Michael Bolin, Daniel
  Cheng, Negar Ghorbani, Renuka Fernandez, and Nachiappan Nagappan.
\newblock Codecompose: A large-scale industrial deployment of ai-assisted code
  authoring.
\newblock {\em arXiv preprint arXiv:2305.12050}, 2023.

\bibitem{kulal2019spoc}
Sumith Kulal, Panupong Pasupat, Kartik Chandra, Mina Lee, Oded Padon, Alex
  Aiken, and Percy~S Liang.
\newblock Spoc: Search-based pseudocode to code.
\newblock {\em Advances in Neural Information Processing Systems}, 32, 2019.

\bibitem{liu2023llm+}
Bo~Liu, Yuqian Jiang, Xiaohan Zhang, Qiang Liu, Shiqi Zhang, Joydeep Biswas,
  and Peter Stone.
\newblock Llm+ p: Empowering large language models with optimal planning
  proficiency.
\newblock {\em arXiv preprint arXiv:2304.11477}, 2023.

\bibitem{weiss2016survey}
Karl Weiss, Taghi~M Khoshgoftaar, and DingDing Wang.
\newblock A survey of transfer learning.
\newblock {\em Journal of Big data}, 3(1):1--40, 2016.

\bibitem{lester2021power}
Brian Lester, Rami Al-Rfou, and Noah Constant.
\newblock The power of scale for parameter-efficient prompt tuning.
\newblock {\em arXiv preprint arXiv:2104.08691}, 2021.

\bibitem{dettmers2023qlora}
Tim Dettmers, Artidoro Pagnoni, Ari Holtzman, and Luke Zettlemoyer.
\newblock Qlora: Efficient finetuning of quantized llms.
\newblock {\em arXiv preprint arXiv:2305.14314}, 2023.

\bibitem{dong2022survey}
Qingxiu Dong, Lei Li, Damai Dai, Ce~Zheng, Zhiyong Wu, Baobao Chang, Xu~Sun,
  Jingjing Xu, and Zhifang Sui.
\newblock A survey for in-context learning.
\newblock {\em arXiv preprint arXiv:2301.00234}, 2022.

\bibitem{gptPlugin}
OpenAI.
\newblock {ChatGPT plugins}, 2023.

\bibitem{gptFuncCall}
OpenAI.
\newblock {Function calling and other API updates}, 2023.

\bibitem{noauthor_langchain_nodate}
LangChian.
\newblock Introduction | langchain, 2023.

\bibitem{noauthor_auto-gpt_2023}
Auto-{GPT}: {An} {Autonomous} {GPT}-4 {Experiment}, June 2023.
\newblock original-date: 2023-03-16T09:21:07Z.

\bibitem{noauthor_babyagi_nodate}
BabyAGI.
\newblock {Translations: | BabyAGI}, 2023.

\bibitem{mckenna2023sources}
Nick McKenna, Tianyi Li, Liang Cheng, Mohammad~Javad Hosseini, Mark Johnson,
  and Mark Steedman.
\newblock Sources of hallucination by large language models on inference tasks.
\newblock {\em arXiv preprint arXiv:2305.14552}, 2023.

\bibitem{manakul2023selfcheckgpt}
Potsawee Manakul, Adian Liusie, and Mark~JF Gales.
\newblock Selfcheckgpt: Zero-resource black-box hallucination detection for
  generative large language models.
\newblock {\em arXiv preprint arXiv:2303.08896}, 2023.

\bibitem{peng2023check}
Baolin Peng, Michel Galley, Pengcheng He, Hao Cheng, Yujia Xie, Yu~Hu, Qiuyuan
  Huang, Lars Liden, Zhou Yu, Weizhu Chen, et~al.
\newblock Check your facts and try again: Improving large language models with
  external knowledge and automated feedback.
\newblock {\em arXiv preprint arXiv:2302.12813}, 2023.

\bibitem{sun2022contrastive}
Weiwei Sun, Zhengliang Shi, Shen Gao, Pengjie Ren, Maarten de~Rijke, and
  Zhaochun Ren.
\newblock Contrastive learning reduces hallucination in conversations.
\newblock {\em arXiv preprint arXiv:2212.10400}, 2022.

\bibitem{li2023skcoder}
Jia Li, Yongmin Li, Ge~Li, Zhi Jin, Yiyang Hao, and Xing Hu.
\newblock Skcoder: A sketch-based approach for automatic code generation.
\newblock {\em arXiv preprint arXiv:2302.06144}, 2023.

\bibitem{jiang2023self}
Xue Jiang, Yihong Dong, Lecheng Wang, Qiwei Shang, and Ge~Li.
\newblock Self-planning code generation with large language model.
\newblock {\em arXiv preprint arXiv:2303.06689}, 2023.

\bibitem{wang2022compilable}
Xin Wang, Yasheng Wang, Yao Wan, Fei Mi, Yitong Li, Pingyi Zhou, Jin Liu, Hao
  Wu, Xin Jiang, and Qun Liu.
\newblock Compilable neural code generation with compiler feedback.
\newblock {\em arXiv preprint arXiv:2203.05132}, 2022.

\bibitem{chen2023improving}
Angelica Chen, Jérémy Scheurer, Tomasz Korbak, Jon~Ander Campos, Jun~Shern
  Chan, Samuel~R. Bowman, Kyunghyun Cho, and Ethan Perez.
\newblock Improving code generation by training with natural language feedback,
  2023.

\bibitem{bixby2007gurobi}
Bob Bixby.
\newblock The gurobi optimizer.
\newblock {\em Transp. Research Part B}, 41(2):159--178, 2007.

\bibitem{chen2021evaluating}
Mark Chen, Jerry Tworek, Heewoo Jun, Qiming Yuan, Henrique Ponde de~Oliveira
  Pinto, Jared Kaplan, Harri Edwards, Yuri Burda, Nicholas Joseph, Greg
  Brockman, et~al.
\newblock Evaluating large language models trained on code.
\newblock {\em arXiv preprint arXiv:2107.03374}, 2021.

\bibitem{zhou2022domain}
Kaiyang Zhou, Ziwei Liu, Yu~Qiao, Tao Xiang, and Chen~Change Loy.
\newblock Domain generalization: A survey.
\newblock {\em IEEE Transactions on Pattern Analysis and Machine Intelligence},
  2022.

\bibitem{pennington2014glove}
Jeffrey Pennington, Richard Socher, and Christopher~D Manning.
\newblock Glove: Global vectors for word representation.
\newblock In {\em Proceedings of the 2014 conference on empirical methods in
  natural language processing (EMNLP)}, pages 1532--1543, 2014.

\bibitem{cai2023large}
Tianle Cai, Xuezhi Wang, Tengyu Ma, Xinyun Chen, and Denny Zhou.
\newblock Large language models as tool makers.
\newblock {\em arXiv preprint arXiv:2305.17126}, 2023.

\bibitem{wiegreffe2021reframing}
Sarah Wiegreffe, Jack Hessel, Swabha Swayamdipta, Mark Riedl, and Yejin Choi.
\newblock Reframing human-ai collaboration for generating free-text
  explanations.
\newblock {\em arXiv preprint arXiv:2112.08674}, 2021.

\bibitem{gunasekar2023textbooks}
Suriya Gunasekar, Yi~Zhang, Jyoti Aneja, Caio C{\'e}sar~Teodoro Mendes, Allie
  Del~Giorno, Sivakanth Gopi, Mojan Javaheripi, Piero Kauffmann, Gustavo
  de~Rosa, Olli Saarikivi, et~al.
\newblock Textbooks are all you need.
\newblock {\em arXiv preprint arXiv:2306.11644}, 2023.

\end{thebibliography}
\bibliographystyle{unsrt}


\clearpage 

\appendix

\section{Intelligent Fulfillment System}
\label{sec:app:ifs}

In this section, we present a partial formulation of the optimization in the Intelligent Fulfillment System that assigns and ships servers from the warehouse to the data centers.

\subsection{Main Decisions}


We introduce the following variables:

\begin{itemize}
\item $z_{dt} \in \{0,1\}$: equals 1 if demand $d$ docks on day $t$, and 0 otherwise
\item $u_{dr} \in \{0,1\}$: equals 1 if demand $d$ docks on row $r$, and 0 otherwise
\item $w_{ds} \in \{0,1\}$: equals 1 if $d$ is fulfilled using supplier $s$, and 0 otherwise
\item $y_{d,dc,t} \in \{0,1\}$: equals 1 if $d$ docks at datacenter $dc$ on day $t$, and 0 otherwise.
\item $v_{d,s,t} \geq 0$ : whether demand $d$ docks on day $t$ using supplier $s$ or not 
\end{itemize}

\subsection{Constraints}

This section describes some of the constraints in the formulation. 

\paragraph{Docking day.} The docking for each demand takes place on a single day.
\begin{equation*}
\sum_{t} z_{dt} \leq 1 \quad \quad \forall d
\end{equation*}

\paragraph{Datacenter dockings.} For each demand $d$, we dock at a datacenter $dc$ on a specific day $t$ only if the selected row belongs to that datacenter $dc$ and the selected day is that particular day $t$.  
\begin{equation*}
\sum_{dc} y_{d,dc,t} \leq z_{dt} \quad \quad \forall d, t
\end{equation*}
\begin{equation*}
\sum_t y_{d,dc,t} = \sum_{r \in rows(dc)} u_{dr} \quad \quad \forall d,dc
\end{equation*}

\paragraph{Datacenters' daily capacities.} There are restrictions $restr$ on the daily amount of dockings that sets of datacenters can handle. Let $R_d$ denote the number of racks required for demand $d$.
\begin{equation*}
\sum_{d, dc \in DC(restr)} y_{d,dc,t} \cdot R_d \leq \mathrm{DockRestrAvailCap}(restr,t) \quad \forall restr \in Restrictions, t
\end{equation*}

\paragraph{Single supplier.} Each demand must be fulfilled by a single supplier. A row is selected for a demand only if a supplier has been found.
\begin{equation*}
\sum_{s} w_{ds} \leq 1 \quad \quad \forall d
\end{equation*}
\begin{equation*}
u_{dr} \leq \sum_{s} w_{ds} \quad \quad \forall d,r
\end{equation*}

\paragraph{Auxiliary supplier variables.} Connecting variables $v_{dst}$ with the rest of the variables. 

\begin{equation*}
z_{dt} = \sum_s v_{dst} \quad \quad \forall d,t
\end{equation*}

\begin{equation*}
w_{ds} = \sum_t v_{dst} \quad \quad \forall d,t
\end{equation*}


\paragraph{Supply availability.} We have a set of supply pools with a certain capacity (amount of available supply) evaluated at times $ct$. We need to make sure that the supply $s$ we consume from each supply pool $sp$ is available at the time $t$ that we consume it. The time where each supply becomes available depends on its lead time. 

\begin{equation*}
\sum_{d, s \in sp, t \leq leadtime(ct, d, s)} v_{dst} \leq \mathrm{Available\_Supply}(sp,ct) \quad \quad \forall sp, ct
\end{equation*}

\paragraph{Overrides.} Some demand-supply combinations might be undesirable or disallowed for some reason. These can be explicitly blocked. Let $B$ denote the set of blocked pairs.
\begin{equation*}
w_{ds} = 0  \quad \quad \forall (d,s) \in B
\end{equation*}

\subsection{Objective} 

Our goal is to minimize the total cost which is the aggregate of multiple components, including the cost of docking too early or too late compared to the ideal dock-date of each demand, the cost of not fulfilling demands, and the shipping cost, among others.

\begin{equation*}
\mathrm{DockCost} = \sum_{d,t} z_{dt} \cdot \mathrm{Demand\_Day\_DockCost}(d,t) 
\end{equation*}

\begin{equation*}
\mathrm{NoDockCost} = \sum_{d} (1-\sum_t z_{dt}) \cdot \mathrm{Unsatisfied\_Cost}(d)
\end{equation*}

\begin{equation*}
\mathrm{ShippingCost} = \sum_{d,s} w_{ds} \cdot \mathrm{Transit\_Ship\_Cost}(d,s) 
\end{equation*}

\section{Engineering Details}
\label{sec:engineer_tricks}

\begin{figure}[h!]
    \centering
    \includegraphics[width=0.70\textwidth]{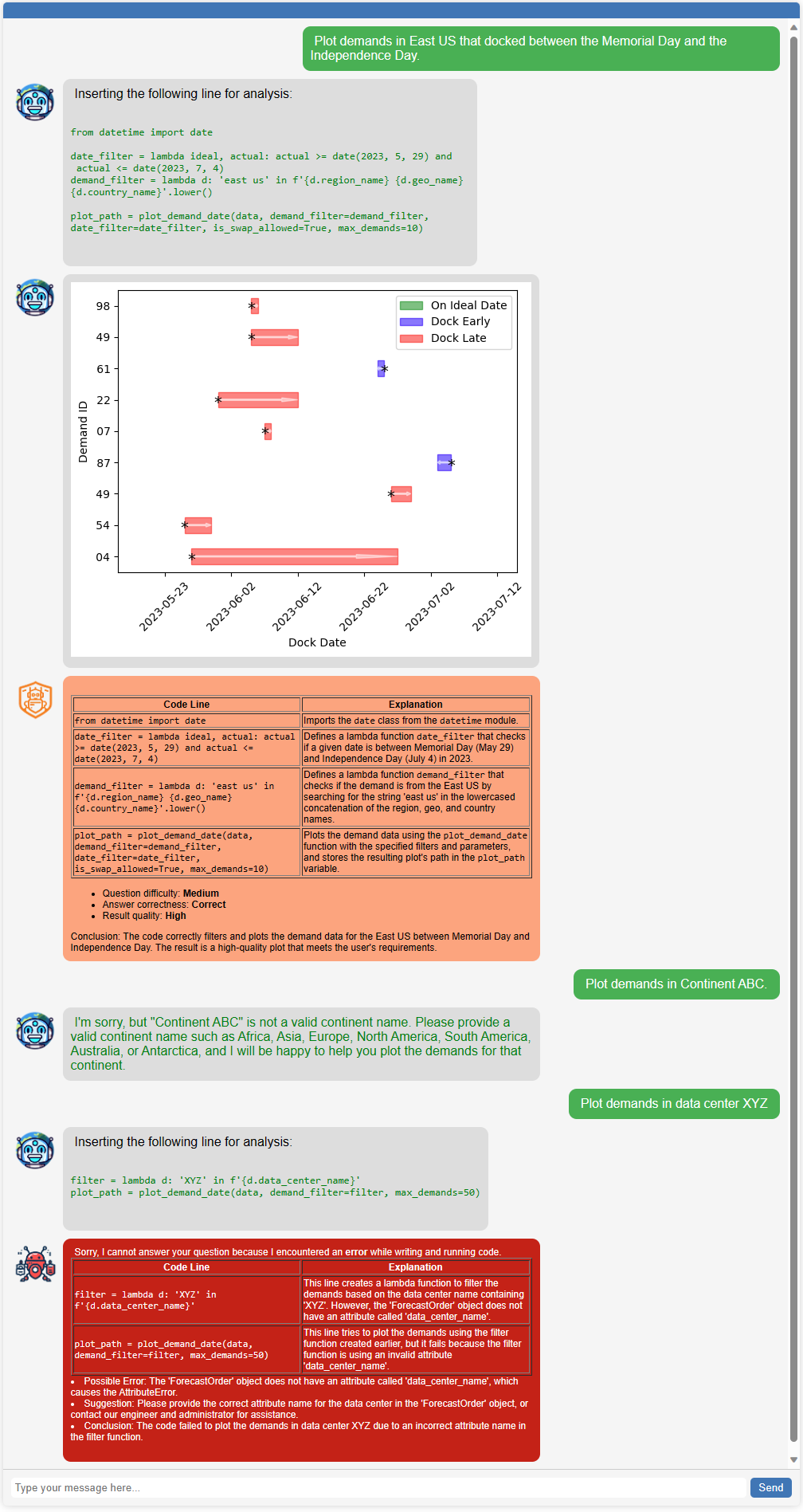}
    \caption{\name{} for \azure{} IFS. Intermediate results from agents are shown in the screenshot.}
    \label{fig:screenshot-details}
\end{figure}

Figure \ref{fig:screenshot-details}, at the end of this document, presents a detailed screenshot of \name{} with \azure{} IFS, including intermediate results for illustration purposes.

\subsection{Useful Tricks}

\paragraph{SQL}: Many LLMs are trained with SQL database. Hence, saving optimization input and output data into SQL could make the system easier to use and more explainable. 

\paragraph{Logical simplification:}
If the prompt is not designed well, the LLM might make many simple logical mistakes (e.g., ``not use" v.s. ``use", before v.s. after, etc.).

\paragraph{Intermediate outputs.} When dealing with complex prompts, providing intermediate outputs can help keep the LLM on track. By returning intermediate results or steps, the LLM can check the consistency of its process, making it easier to debug and refine.

\subsection{Failed Attempts}

\paragraph{Chain of thought (CoT) failures.} Unlike many recent studies \cite{cai2023large} that have found that LLMs have strong CoT abilities, we found CoT is not helpful for writing complex code. This is another reason why we integrated the helper functions in the application-specific tools, which outperformed CoT. Our hypothesis is that if the LLM makes one mistake in the thinking chain, then the whole response would be wrong because correcting its own mistakes is hard. 

\paragraph{Overuse of prompt engineering:} While prompt engineering can often lead to improved results, overdoing it can sometimes lead to worse outcomes. When the prompts become too complex or too specific, the LLM might not understand them correctly or might overfit to the specific prompt structure, limiting its ability to handle a variety of questions.

\section{Coffee Distribution Example}
\label{sec:coffee_code}

\subsection{Code}

\begin{verbatim}
import time

from gurobipy import GRB, Model

# Example data

capacity_in_supplier = {'supplier1': 150, 'supplier2': 50, 'supplier3': 100}

shipping_cost_from_supplier_to_roastery = {
    ('supplier1', 'roastery1'): 5,
    ('supplier1', 'roastery2'): 4,
    ('supplier2', 'roastery1'): 6,
    ('supplier2', 'roastery2'): 3,
    ('supplier3', 'roastery1'): 2,
    ('supplier3', 'roastery2'): 7
}

roasting_cost_light = {'roastery1': 3, 'roastery2': 5}

roasting_cost_dark = {'roastery1': 5, 'roastery2': 6}

shipping_cost_from_roastery_to_cafe = {
    ('roastery1', 'cafe1'): 5,
    ('roastery1', 'cafe2'): 3,
    ('roastery1', 'cafe3'): 6,
    ('roastery2', 'cafe1'): 4,
    ('roastery2', 'cafe2'): 5,
    ('roastery2', 'cafe3'): 2
}

light_coffee_needed_for_cafe = {'cafe1': 20, 'cafe2': 30, 'cafe3': 40}

dark_coffee_needed_for_cafe = {'cafe1': 20, 'cafe2': 20, 'cafe3': 100}

cafes = list(set(i[1] for i in shipping_cost_from_roastery_to_cafe.keys()))
roasteries = list(
    set(i[1] for i in shipping_cost_from_supplier_to_roastery.keys()))
suppliers = list(
    set(i[0] for i in shipping_cost_from_supplier_to_roastery.keys()))

# OPTIGUIDE DATA CODE GOES HERE

# Create a new model
model = Model("coffee_distribution")

# Create variables
x = model.addVars(shipping_cost_from_supplier_to_roastery.keys(),
                  vtype=GRB.INTEGER,
                  name="x")
y_light = model.addVars(shipping_cost_from_roastery_to_cafe.keys(),
                        vtype=GRB.INTEGER,
                        name="y_light")
y_dark = model.addVars(shipping_cost_from_roastery_to_cafe.keys(),
                       vtype=GRB.INTEGER,
                       name="y_dark")

# Set objective
model.setObjective(
    sum(x[i] * shipping_cost_from_supplier_to_roastery[i]
        for i in shipping_cost_from_supplier_to_roastery.keys()) +
    sum(roasting_cost_light[r] * y_light[r, c] +
        roasting_cost_dark[r] * y_dark[r, c]
        for r, c in shipping_cost_from_roastery_to_cafe.keys()) + sum(
            (y_light[j] + y_dark[j]) * shipping_cost_from_roastery_to_cafe[j]
            for j in shipping_cost_from_roastery_to_cafe.keys()), GRB.MINIMIZE)

# Conservation of flow constraint
for r in set(i[1] for i in shipping_cost_from_supplier_to_roastery.keys()):
    model.addConstr(
        sum(x[i]
            for i in shipping_cost_from_supplier_to_roastery.keys()
            if i[1] == r) == sum(
                y_light[j] + y_dark[j]
                for j in shipping_cost_from_roastery_to_cafe.keys()
                if j[0] == r), f"flow_{r}")

# Add supply constraints
for s in set(i[0] for i in shipping_cost_from_supplier_to_roastery.keys()):
    model.addConstr(
        sum(x[i]
            for i in shipping_cost_from_supplier_to_roastery.keys()
            if i[0] == s) <= capacity_in_supplier[s], f"supply_{s}")

# Add demand constraints
for c in set(i[1] for i in shipping_cost_from_roastery_to_cafe.keys()):
    model.addConstr(
        sum(y_light[j]
            for j in shipping_cost_from_roastery_to_cafe.keys()
            if j[1] == c) >= light_coffee_needed_for_cafe[c],
        f"light_demand_{c}")
    model.addConstr(
        sum(y_dark[j]
            for j in shipping_cost_from_roastery_to_cafe.keys()
            if j[1] == c) >= dark_coffee_needed_for_cafe[c], f"dark_demand_{c}")

# Optimize model
model.optimize()
m = model

# OPTIGUIDE CONSTRAINT CODE GOES HERE

# Solve
m.update()
model.optimize()

print(time.ctime())
if m.status == GRB.OPTIMAL:
    print(f'Optimal cost: {m.objVal}')
else:
    print("Not solved to optimality. Optimization status:", m.status)
\end{verbatim}

\subsection{Question and Ground Truth Macros}

\begin{verbatim}
QUESTION:
What would happen if demand at cafe {{VALUE-CAFE}} increased by {{VALUE-NUMBER}}%?
VALUE-CAFE: random.choice(cafes)
VALUE-NUMBER: random.randrange(5,30)
DATA CODE:
light_coffee_needed_for_cafe[{{VALUE-CAFE}}] =  \
    light_coffee_needed_for_cafe[{{VALUE-CAFE}}] * (1 + {{VALUE-NUMBER}}/100)
dark_coffee_needed_for_cafe[{{VALUE-CAFE}}] =  \
    dark_coffee_needed_for_cafe[{{VALUE-CAFE}}] * (1 + {{VALUE-NUMBER}}/100)
TYPE: demand-increase

QUESTION:
What if demand for light coffee at cafe {{VALUE-CAFE}} increased by {{VALUE-NUMBER}}%?
VALUE-CAFE: random.choice(cafes)
VALUE-NUMBER: random.randrange(5,30)
DATA CODE:
light_coffee_needed_for_cafe[{{VALUE-CAFE}}] = \
    light_coffee_needed_for_cafe[{{VALUE-CAFE}}] * (1 + {{VALUE-NUMBER}}/100)
TYPE: demand-increase-light

QUESTION:
What would happen if the demand at all cafes doubled?
DATA CODE:
for c in cafes:
    light_coffee_needed_for_cafe[c] = light_coffee_needed_for_cafe[c] * 2
    dark_coffee_needed_for_cafe[c] = dark_coffee_needed_for_cafe[c] * 2
TYPE: demand-increase-all

QUESTION:
Why are we using supplier {{VALUE-SUPPLIER}} for roasting facility {{VALUE-ROASTERY}}?
VALUE-SHIPPINGS: [(s, r) for (s, r), value in x.items() if value.X >= 0.999]
VALUE-IDX: random.randint(0, len({{VALUE-SHIPPINGS}}) - 1)
VALUE-SUPPLIER: {{VALUE-SHIPPINGS}}[{{VALUE-IDX}}][0]
VALUE-ROASTERY: {{VALUE-SHIPPINGS}}[{{VALUE-IDX}}][1]
CONSTRAINT CODE:
m.addConstr(x[{{VALUE-SUPPLIER}},{{VALUE-ROASTERY}}] == 0, "_")
TYPE: supply-roastery

QUESTION: 
Assume cafe {{VALUE-CAFE}} can exclusively buy coffee from roasting facility
{{VALUE-ROASTERY}}, and conversely, roasting facility {{VALUE-ROASTERY}} 
can only sell its coffee to cafe {{VALUE-CAFE}}. How does that affect the outcome?
VALUE-ROASTERY: random.choice(roasteries)
VALUE-CAFE: random.choice(cafes)
CONSTRAINT CODE:
for c in cafes:
    if c != {{VALUE-CAFE}}:
        m.addConstr(y_light[{{VALUE-ROASTERY}}, c] == 0, "_")
        m.addConstr(y_dark[{{VALUE-ROASTERY}}, c] == 0, "_")
for r in roasteries:
    if r != {{VALUE-ROASTERY}}:
        m.addConstr(y_light[r,{{VALUE-CAFE}}] == 0, "_")
        m.addConstr(y_dark[r,{{VALUE-CAFE}}] == 0, "_")
TYPE: exclusive-roastery-cafe

QUESTION:
What if roasting facility {{VALUE-ROASTERY}} can only be used for cafe {{VALUE-CAFE}}?
VALUE-ROASTERY: random.choice(roasteries)
VALUE-CAFE: random.choice(cafes)
CONSTRAINT CODE:
for c in cafes:
    if c != {{VALUE-CAFE}}:
        m.addConstr(y_light[{{VALUE-ROASTERY}}, c] == 0, "_")
        m.addConstr(y_dark[{{VALUE-ROASTERY}}, c] == 0, "_")
TYPE: incompatible-roastery-cafes

QUESTION:
What if supplier {{VALUE-SUPPLIER}} can now provide only half of the quantity?
VALUE-SUPPLIER: random.choice(suppliers)
DATA CODE:
capacity_in_supplier[{{VALUE-SUPPLIER}}] = capacity_in_supplier[{{VALUE-SUPPLIER}}]/2
TYPE: supplier-capacity

QUESTION:
The per-unit cost from supplier {{VALUE-SUPPLIER}} to roasting facility {{VALUE-ROASTERY}} 
is now {{VALUE-NUMBER}}. How does that affect the total cost?
VALUE-SUPPLIER: random.choice(suppliers)
VALUE-ROASTERY: random.choice(roasteries)
VALUE-NUMBER: random.randrange(1,10)
DATA CODE:
shipping_cost_from_supplier_to_roastery[{{VALUE-SUPPLIER}},{{VALUE-ROASTERY}}] = \
    {{VALUE-NUMBER}}
TYPE: supplier-roastery-shipping

QUESTION:
What would happen if roastery 2 produced at least as much light coffee as roastery 1?
CONSTRAINT CODE:
m.addConstr(sum(y_light['roastery1',c] for c in cafes) 
            <= sum(y_light['roastery2',c] for c in cafes), "_")
TYPE: light-quantities-roasteries

QUESTION:
What would happen if roastery 1 produced less light coffee than roastery 2?
CONSTRAINT CODE:
m.addConstr(sum(y_light['roastery1',c] for c in cafes) 
            <= sum(y_light['roastery2',c] for c in cafes) - 1, "_")
TYPE: light-quantities-roasteries

QUESTION:
What will happen if supplier 1 ships more to roastery 1 than roastery 2?
CONSTRAINT CODE:
m.addConstr(x['supplier1','roastery1'] >= x['supplier1','roastery2'] + 1, "_")
TYPE: shipping-quantities-roasteries

QUESTION:
What will happen if supplier 1 ships to roastery 1 at least as much as to roastery 2?
CONSTRAINT CODE:
m.addConstr(x['supplier1','roastery1'] >= x['supplier1','roastery2'], "_")
TYPE: shipping-quantities-roasteries

QUESTION:
Why not only use a single supplier for roastery 2?
CONSTRAINT CODE:
z = m.addVars(suppliers, vtype=GRB.BINARY, name="z")
m.addConstr(sum(z[s] for s in suppliers) <= 1, "_")
for s in suppliers:
    m.addConstr(x[s,'roastery2'] <= capacity_in_supplier[s] * z[s], "_")
TYPE: single-supplier-roastery
\end{verbatim}


\begin{figure}[h!]
    \centering
\begin{tikzpicture}
\begin{axis}[
  xlabel={Number of Shots (learning examples)},
  ylabel={Coding Accuracy $\%$},
  xmin=0, xmax=10,
  ymin=55, ymax=93,
  grid=both,
  minor tick num=2
]
\addplot[msblue, opacity=0.5, line width=3, mark=*] coordinates {(0, 59) (1, 66) (3, 69) (5,73) (10,80)};
\addlegendentry{Nearest Neighbours}
\addplot[msred, opacity=0.5, line width=3, mark=square*] coordinates {(0, 59) (1, 66) (3, 74) (5, 78) (10, 84)};
\addlegendentry{Random Choices}
\end{axis}
\end{tikzpicture}
    \caption{Out-of-distribution evaluation for GPT-4. We compare the different training example selection methods here.}
    \label{fig:ood-embedding}
\end{figure}

\end{document}